\documentclass[sigconf]{acmart}
\usepackage{lineno,hyperref}
\usepackage{colortbl}
\usepackage{graphicx}
\usepackage{float}

\usepackage{latexsym}
\usepackage{url}

\usepackage{xspace}
\usepackage{graphicx}
\usepackage{multirow}
\usepackage{makecell}
\usepackage{lscape}
\usepackage{bm}
\usepackage{algorithm}
\usepackage{algpseudocode}
\usepackage{framed}
\usepackage{color}
\usepackage{xcolor}
\usepackage{booktabs,colortbl}
\usepackage{bbm}
\usepackage{subfigure}
\usepackage{multirow}

\usepackage{amssymb}
\usepackage{amsmath}
\usepackage{multirow}
\usepackage{makecell}
\usepackage{tabularx}
\usepackage{booktabs}
\usepackage{setspace}
\usepackage{pifont}
\usepackage{arydshln}
\definecolor{lightgrayv}{HTML}{F4F3F8} 
\definecolor{grayv}{HTML}{707070}

\newcommand{\eg}{\emph{e.g.,}\xspace}
\newcommand{\ie}{\emph{i.e.,}\xspace}

\newcommand{\baby}{\textsc{DaedCmd}\xspace}
\AtBeginDocument{%
  }

\setcopyright{acmlicensed}
\copyrightyear{2025}
\acmYear{2025}
\acmDOI{XXXXXXX.XXXXXXX}

\acmConference[MM '25] {Proceedings of the 33nd ACM International Conference on Multimedia}{October 27--31, 2025}{Dublin, Ireland}
\acmBooktitle{Proceedings of the 33nd ACM International Conference on Multimedia (MM '25), October 27--31, 2025, Dublin, Ireland}
\acmISBN{978-1-4503-XXXX-X/18/06}

\begin{document}

\title{\textit{Remember Past, Anticipate Future}: Learning Continual Multimodal Misinformation Detectors}

\author{Bing Wang}
\orcid{0000-0002-1304-3718}
\affiliation{
  \institution{College of Computer Science and Technology, Jilin University}
  \city{Changchun}
  \state{Jilin}
  \country{China}}
\email{wangbing1416@gmail.com}

\author{Ximing Li}
\thanks{* Ximing Li is the corresponding author. Bing Wang, Ximing Li, Mengzhe Ye, and Changchun Li are also affiliated with Key Laboratory of Symbolic Computation and Knowledge Engineering of the Ministry of Education, Jilin University.}
\authornotemark[1]
\orcid{0000-0001-8190-5087}
\affiliation{
  \institution{College of Computer Science and Technology, Jilin University}
  \city{Changchun}
  \state{Jilin}
  \country{China}}
\email{liximing86@gmail.com}

\author{Mengzhe Ye}
\orcid{0009-0001-2949-4134}
\affiliation{
  \institution{College of Software, Jilin University}
  \city{Changchun}
  \state{Jilin}
  \country{China}}
\email{yemengzhe1@gmail.com}

\author{Changchun Li}
\orcid{0000-0002-8001-2655}
\affiliation{
  \institution{College of Computer Science and Technology, Jilin University}
  \city{Changchun}
  \state{Jilin}
  \country{China}}
\email{changchunli93@gmail.com}

\author{Bo Fu}
\orcid{0000-0001-7030-821X}
\affiliation{
  \institution{School of Computer and  Artificial Intelligence, Liaoning Normal University}
  \city{Dalian}
  \state{Liaoning}
  \country{China}}
\email{fubo@lnnu.edu.cn}

\author{Jianfeng Qu}
\orcid{0000-0002-3212-7618}
\affiliation{
  \institution{School of Computer Science and Technology, Soochow University}
  \city{Suzhou}
  \state{Jiangsu}
  \country{China}}
\email{jfqu@suda.edu.cn}

\author{Lin Yuanbo Wu}
\orcid{0000-0001-6119-058X}
\affiliation{
  \institution{Department of Computer Science, Swansea University}
  \city{Swansea}
  \state{Wales}
  \country{United Kingdom}}
\email{l.y.wu@swansea.ac.uk}



\begin{abstract}
  Nowadays, misinformation articles, especially multimodal ones, are widely spread on social media platforms and cause serious negative effects. To control their propagation, \textbf{M}ultimodal \textbf{M}isinformation \textbf{D}etection (\textbf{MMD}) becomes an active topic in the community to automatically identify misinformation. Previous MMD methods focus on supervising detectors by collecting offline data. However, in real-world scenarios, new events always continually emerge, making MMD models trained on offline data consistently outdated and ineffective. To address this issue, training MMD models under online data streams is an alternative, inducing an emerging task named \textbf{continual MMD}. 
Unfortunately, it is hindered by two major challenges. First, training on new data consistently decreases the detection performance on past data, named past knowledge forgetting. Second, the social environment constantly evolves over time, affecting the generalization on future data.
To alleviate these challenges, we propose to remember past knowledge by isolating interference between event-specific parameters with a Dirichlet process-based mixture-of-expert structure, and anticipate future environmental distributions by learning a continuous-time dynamics model. Accordingly, we induce a new continual MMD method \baby. Extensive experiments demonstrate that \baby can consistently and significantly outperform the compared methods, including six MMD baselines and three continual learning methods.

\end{abstract}




\maketitle

\section{Introduction}
Currently, due to the widespread popularity of various social media platforms, \eg X.com and Reddit, users are able to conveniently post multimedia content, thereby connecting with people around the world. However, these platforms have also created a breeding ground for the intentional spread of misinformation, especially multimodal ones, thereby causing serious threats to people's mental and financial security \citep{vicario2016spreading}. 
To control the spread of misinformation, the initial task is to automatically detect it within a vast number of social media posts. Accordingly, the community resorts to an active topic named \textbf{M}ultimodal \textbf{M}isinformation \textbf{D}etection (\textbf{MMD}).

\begin{figure}[t]
  \centering
  \includegraphics[scale=0.46]{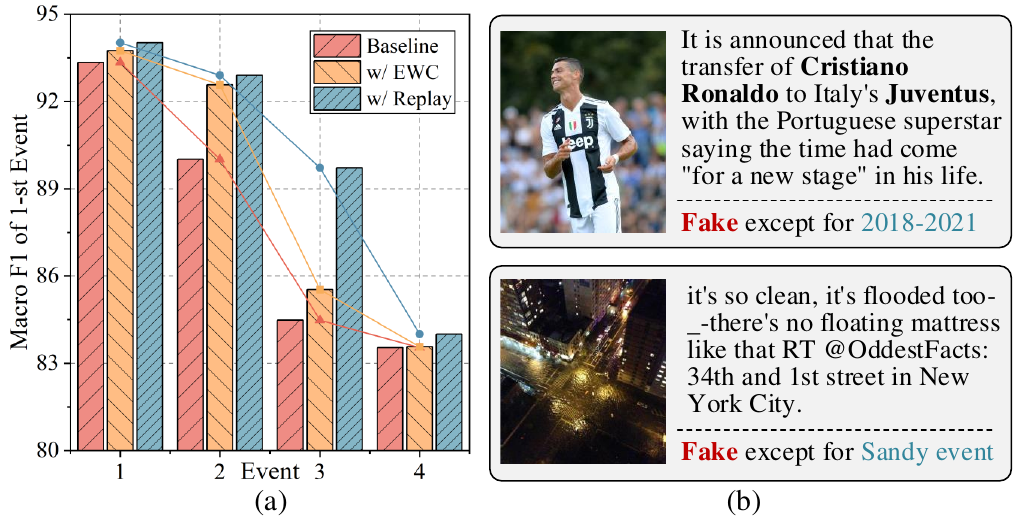}
  \caption{Preliminary experiments that demonstrate Macro F1 scores of the first training event on \textit{Weibo} when sequentially training an MMD model on four events, and representative cases that depict the dynamic social environment.}
  \label{preliminary}
\end{figure}

Previous MMD methods develop various deep models to map multimodal articles to a hidden semantic space and learn the potential relationships between these multimodal semantics and veracity labels, \eg real and fake \citep{chen2022cross,ying2023bootstrapping,wang2024why}. For example, \citet{ying2023bootstrapping} learn an improved multi-gate mixture-of-expert to generate multi-view features.
Typically, these methods are solely effective on \textit{offline} data \citep{hu2023learn,ding2025evolvedetector}, where large-scale data is pre-collected for training and evaluating MMD models.
However, in real-world online social media environments, new events always continually emerge, which renders MMD models trained on offline data consistently outdated and ineffective when deployed in practical scenarios. To address this issue, training MMD models under online data streams is an alternative, inducing an emerging task named \textbf{continual MMD}.

Generally, continual MMD involves incrementally training models on different data/events. Unfortunately, it is hindered by two major challenges \textbf{past knowledge forgetting} and \textbf{social environment evolving}. 
First, when facing a series of data streams, especially large-scale data from social media, training on new data consistently causes the detection performance of MMD models to decrease on past data, named knowledge forgetting \citep{kirkpatrick2016overcoming,d2019episodic}. We provide a preliminary experiment in Fig.~\ref{preliminary}(a) to illustrate this phenomenon.
On the other hand, in online scenarios, the social environment consistently changes over time, which leads to a gradual evolution of the data distributions for continual MMD, especially fake ones \citep{hu2023learn,zhang2024evolving}. For example, as depicted in Fig.~\ref{preliminary}(b), the arrival of a new event may lead to a change in the veracity label of an article.
Accordingly, this evolving social environment also affects the generalization of MMD models on future data.

To alleviate these challenges, we propose to remember the past knowledge by isolating interference between event-specific parameters, and anticipate the future social environment by learning a continuous-time dynamics model of the environmental distribution. 
Accordingly, we induce a new continual MMD framework, namely \textbf{D}ynamically \textbf{A}daptive \textbf{E}xperts and \textbf{D}istributions for \textbf{C}ontinual \textbf{MMD} (\textbf{\baby}). Specifically, the basic idea of \baby is two-folds. First, to address the knowledge forgetting, we design a Mixture-of-Expert (MoE) structure, which includes event-shared experts to store shared knowledge across events, and event-specific experts to isolate model parameters between different events, thus reducing interference between the training gradients of events. Unfortunately, when dealing with a large number of events, constructing an event-specific expert for each event is extremely time-consuming. To this end, we propose a Dirichlet process mixture-based approach to dynamically expand the number of experts, where a new event-specific expert is created only when the data discrepancy between a new event and previous ones exceeds a certain threshold. 
Second, to learn the dynamics of the social environment, we first construct the environment as a Gaussian distribution and use it to learn a continuous-time dynamics model. For future events, we utilize this dynamics model to predict their environmental distribution and sample a new feature from the predicted one for veracity prediction.

We evaluate our method \baby across three continual MMD datasets and compare it with six MMD baselines and three prevalent continual learning methods. Extensive experiments demonstrate that \baby can consistently and significantly outperform these compared methods, and effectively address the present challenges. \textit{Our source code and data are released in \url{https://github.com/wangbing1416/DAEDCMD}.}

Our contributions can be summarized as the following threefolds:
\begin{itemize}
    \item We identify two primary challenges past knowledge forgetting and social environment dynamics when MMD models learn from continual data streams, and propose a new continual MMD method \baby.
    \item To address two challenges, we respectively propose a Dirichlet-expanded MoE structure and a continuous-time dynamics model to learn the distribution dynamics.
    \item Extensive experiments are conducted across three continual MMD datasets to demonstrate that \baby outperforms SOTA MMD baselines and continual learning methods.
\end{itemize}

\section{Related Works}

In this section, we review related literature about multimodal misinformation detection and continual learning.

\subsection{Multimodal Misinformation Detection}

Previous MMD methods train and evaluate models based on offline collected data. These methods typically employ a variety of deep learning techniques to capture semantic information from multimodal content and learn the relationship between these semantics and their corresponding veracity labels \citep{chen2022cross,ying2023bootstrapping,wang2024harmfully,zhang2024reinforced,wang2024escaping}. 
For example, CAFE \citep{chen2022cross} and COOLANT \citep{wang2023cross} detect misinformation by learning inconsistencies between text and image pairs.
In addition, partial MMD works concentrate on training models with online data, utilizing models trained on past events to detect the veracity of data from future events. They typically propose event adaptation strategies aimed at learning event-invariant features to enhance the model's performance on unseen future events \citep{wang2018eann,wang2021multimodal,zhang2024evolving}.

Among these efforts, a handful of works leverage online data streams to train unimodal detection models while mitigating the catastrophic forgetting problem inherent in continual learning \citep{lee2021dynamically,zuo2022continually,shao2024an,ding2025evolvedetector,han2021continual}.
Some of them focus on supervised learning settings, employing prevalent continual learning strategies, \eg EWC \citep{kirkpatrick2016overcoming,han2021continual} and replay \citep{robins1995catastrophic,lee2021dynamically} methods, to address the continual detection task. For example, EvolveDetector \citep{ding2025evolvedetector} proposes a hard attention-based knowledge-storing mechanism to retain historical event knowledge to achieve efficient knowledge transfer. 
Additionally, another study incorporates active learning into the continual learning process to alleviate the annotation burden \citep{shao2024an}.
Although these studies have also investigated the continual misinformation detection task, they primarily concentrate on unimodal data and mainly address the catastrophic forgetting issue, without considering the dynamic evolution of the social environment.

\subsection{Continual Learning}

A continual learning system aims to sequentially train models on a series of tasks while addressing the catastrophic forgetting issue, which refers to the performance degradation on previously learned tasks \citep{french1993catastrophic,ring1997child}. 
Generally, existing continual learning methods can be broadly categorized into three types: 
\textbf{Architecture-based} approaches isolate the parameters for each task and freeze them during the fine-tuning of other tasks to minimize interference between task-specific parameters \citep{dou2024loramoe,ma2024modula,feng2024mixture}. For example, LoRAMoE \citep{dou2024loramoe} designs a multi-expert LoRA architecture to alleviate the forgetting of pre-trained knowledge when large language models are fine-tuned on downstream tasks;
\textbf{Replay-based} approaches dynamically store partial data or training gradients from previous tasks and use this information to perform joint training with subsequent tasks \citep{d2019episodic,huang2024mitigating}. 
\textbf{Optimization-based} approaches mitigate forgetting by designing objectives or optimization strategies that are resistant to forgetting \citep{lopez2017gradient,saha2021gradient,wang2023orthogonal}. For instance, EWC \citep{kirkpatrick2016overcoming} restricts adjustments to weights that are critical for prior tasks, thereby protecting these weights from excessive changes to preserve the model's memory of old tasks. Orthogonal gradient optimization \citep{saha2021gradient,wang2023orthogonal} reduces interference between task optimization directions by ensuring that the training gradients of new tasks remain orthogonal to those of previous tasks.
In this paper, we adopt a parameter isolation-based approach to design event-shared and -specific experts and dynamically expand the experts to alleviate forgetting in continual MMD.

\section{Our Proposed Method}

In this section, we briefly describe the definition of the continual MMD task and present the \baby method in detail.

\vspace{2pt} \noindent
\textbf{Problem definition of continual MMD.} 
Formally, a typical MMD dataset comprises $N$ samples, denoted as $\mathcal{D} = \{\mathcal{X}_i, y_i\}_{i=1}^N$, where $\mathcal{X}_i = (\mathbf{x}_i^t, \mathbf{x}_i^v)$ represents the text content and images of an article, and $y_i \in \{0, 1\}$ is the corresponding veracity label (0/1 means real/fake).
The goal of MMD is to train a detector $\mathcal{F}_{\boldsymbol{\theta}}(\cdot)$ to predict the veracity of an unseen article.
In the continual MMD scenario, the dataset $\mathcal{D}$ is split into $K$ subsets $\{\mathcal{D}_1, \cdots, \mathcal{D}_K\}$ by events, temporal periods, or domains. The detector $\mathcal{F}_{\boldsymbol{\theta}}(\cdot)$ is then trained incrementally across these subsets, and evaluated on future articles.
For clarity, we summarize the important notations and their descriptions in Table~\ref{notation}.

\subsection{Overview of \baby}

\begin{figure*}[t]
  \centering
  \includegraphics[scale=0.76]{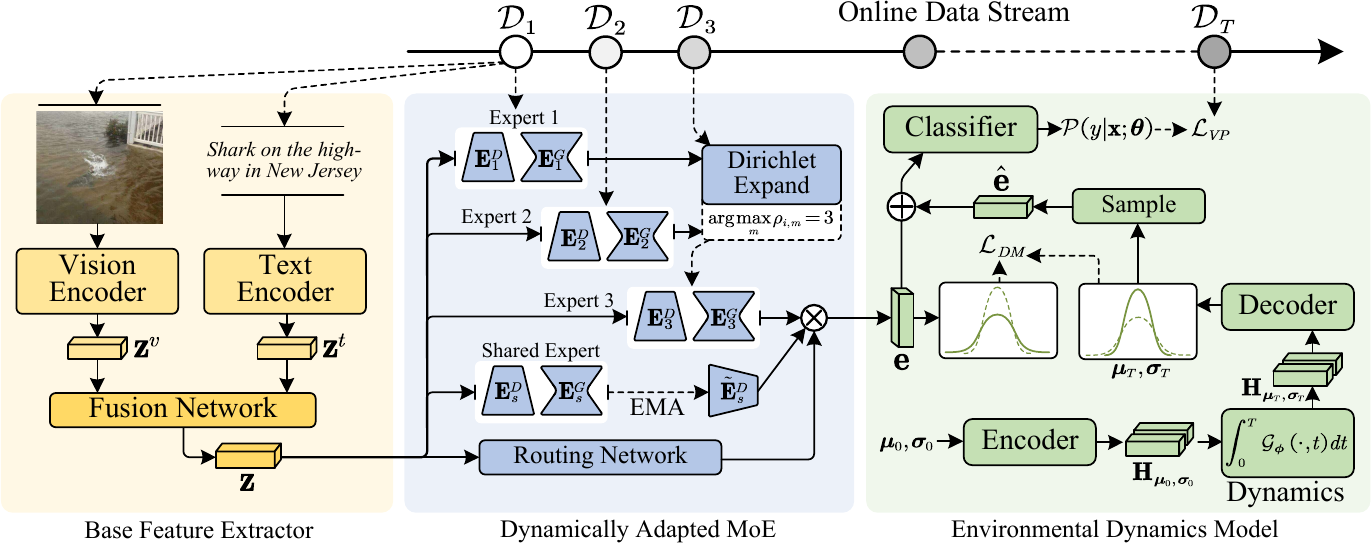}
  \caption{Overall Framework of \baby. Given the text $\mathbf{x}^t$ and image $\mathbf{x}^v$, we first extract their semantics and fuse them as a multimodal feature $\mathbf{z}$. To mitigate forgetting, we propose a dynamically adapted MoE to obtain the forgetting resistant feature $\mathbf{e}$. To learn the dynamically evolved environmental distribution, we construct it as a Gaussian distribution, and learn a dynamics model to predict the environmental distribution of future samples and draw an environmental feature $\mathbf{\hat e}$ for veracity prediction.}
  \label{framework}
\end{figure*}

During the continual training of the MMD model, we observe two key issues. 
First, as the training data, \ie $K$, incrementally expands, the detector's performance on past data consistently declines, referred as knowledge forgetting. Second, on online social media, the social environment in which articles are situated is constantly evolving, making the previously learned experiences of the detector progressively out-of-date.
To address these two challenges, we propose a new framework \baby, which aims to restore the knowledge of past data and simultaneously learn the dynamic environmental distribution. Specifically, \baby consists of three key modules: \textbf{base feature extractor}, \textbf{dynamically adapted MoE}, and \textbf{environmental dynamics model}. For clarity, we depict the overall framework of \baby in Fig.~\ref{framework}. In the following, we briefly present these modules.

\begin{table}[t]
\centering
\renewcommand\arraystretch{1.10}
\small
  \caption{Summary of notations and their descriptions.}
  \label{notation}
  \begin{tabular}{m{2.7cm}<{\centering}|m{4.9cm}<{\centering}}
    \bottomrule
    Notation & Description \\
    \hline
    $\mathcal{D}_k = \{\mathcal{X}_i, y_i\}_{i=1}^{|\mathcal{D}_k|}$ & $k$-th subset of the continual MMD dataset \\
    $\mathbf{z}_i^t, \mathbf{z}_i^v, \mathbf{z}_i$ & text, image and multimodal features \\
    $\mathbf{e}_i, \mathbf{\hat e}_i$ & forgetting resistant/environmental features \\
    $\boldsymbol{\theta} = \{\boldsymbol{\theta}^t, \boldsymbol{\theta}^v, \mathbf{W}_C\}$ & parameter of the continual MMD detector \\
    $\boldsymbol{\phi} = \{\boldsymbol{\phi}_{\boldsymbol{\mu}}, \boldsymbol{\phi}_{\boldsymbol{\sigma}}\}$ & parameter of the dynamics model \\
    $\mathbf{E}_s, \{\mathbf{E}_m\}_{m=1}^M$ & event-shared and event-specific experts \\
    $\mathbb{P}(\mathbf{\hat e}|y=1) = \mathcal{N} \big( \boldsymbol{\mu}, \boldsymbol{\sigma}^2 \big)$ & environmental distribution \\
    $\alpha, \beta, \gamma$ & trade-off hyper-parameters \\
    \bottomrule
  \end{tabular} 
\end{table}

\vspace{2pt} \noindent
\textbf{Base feature extractor.}
This module extracts semantic features of two modality content $\mathbf{x}_i^t$ and $\mathbf{x}_i^v$ and fusing them into a multimodal feature $\mathbf{z}_i$.
Specifically, given a pair of text $\mathbf{x}_i^t$ and an image $\mathbf{x}_i^v$, we use a pre-trained language model, \eg BERT \citep{devlin2019bert}, to extract the text semantic feature $\mathbf{z}_i^t = \mathcal{F}_{\boldsymbol{\theta}^t}(\mathbf{x}_i^t)$, and an image encoder, \eg ResNet \cite{he2016deep} and ViT \citep{dosovitskiy2021an}, to extract the image semantic feature $\mathbf{z}_i^v = \mathcal{F}_{\boldsymbol{\theta}^v}(\mathbf{x}_i^v)$.
To align two unimodal features into a shared feature space, we use a contrastive learning based method \citep{chen2022cross,wang2023cross} as follows:
\begin{equation}
    \label{eq1}
    \mathcal{L}_{CL} = - \frac{1}{N} \sum \nolimits _{i=1}^N \log \frac{e^{\text{sim}(\mathbf{z}_i^t, \mathbf{z}_i^v) / \xi}}
    {\sum \nolimits _{j \in \neg i} e^{\text{sim}(\mathbf{z}_i^t, \mathbf{z}_j^v) / \xi} + e^{\text{sim}(\mathbf{z}_j^t, \mathbf{z}_i^v) / \xi}},
\end{equation}
where $\text{sim}(\cdot\ , \cdot)$ is a similarity measurement function, and $\xi$ is a temperature hyper-parameter.
Then, we directly concatenate two unimodal features and utilize a multi-head self-attention network to fuse them into a multimodal feature $\mathbf{z}_i = [\mathbf{z}_i^t; \mathbf{z}_i^v] \mathbf{W}_A$. In our implementation, the base feature extractor can be replaced by various SOTA MMD methods, \eg BMR \citep{ying2023bootstrapping} and GAMED \citep{shen2025gamed}.

\vspace{2pt} \noindent
\textbf{Dynamically adapted MoE.}
To alleviate the knowledge forgetting issue of continual MMD, we draw inspiration from parameter isolation methods \citep{dou2024loramoe,ma2024modula}, and construct a dynamically adapted Mixture-of-Expert (MoE) structure to obtain a forgetting resistant feature $\mathbf{e}_i = \mathcal{G}_{\boldsymbol{\psi}}(\mathbf{z}_i)$. Specifically, the basic ideas of the dynamically adapted MoE module are two-fold. 
First, existing parameter isolation methods typically create a new expert for each subset, thereby isolating the parameters for each group of data. However, in the real-world large-scale continual MMD scenario, the number of new events keeps increasing, leading to an exponential increase in model complexity.
To address this, we propose to dynamically determine when to create a new expert for a training subset. Accordingly, we employ a Dirichlet process-based approach \citep{lee2020a,ye2021lifelong} to decide whether a new expert should be created for a new subset by optimizing a variational generator with an objective $\mathcal{L}_{VG}$.
Second, to preserve the shared knowledge across data, we divide the experts in the MoE model into event-shared and event-specific ones.

\vspace{2pt} \noindent
\textbf{Environmental dynamics model.}
This module learns the dynamics of the social environmental distribution of misinformation, and uses the predicted environmental distribution as additional feature $\mathbf{\hat e}_i$ for MMD.
Specifically, we first use the multimodal feature $\mathbf{e}_{y=1}$ of fake articles to build a environmental distribution specified by a Gaussian distribution $\mathbb{P}(\mathbf{\hat e}|y=1) = \mathcal{N} \big( \boldsymbol{\mu}, \boldsymbol{\sigma}^2 \big)$, where $\boldsymbol{\mu}$ and $\boldsymbol{\sigma}$ represent the mean and variance, respectively.
Accordingly, to learn the dynamics of the environmental distribution, we train a neural dynamics model $\mathcal{G}_{\boldsymbol{\phi}}(\cdot)$ with an objective $\mathcal{L}_{DM}$ to predict the environmental distribution of future data. Finally, we sample an environmental feature $\mathbf{\hat e}_i$ from these predicted distributions.

Based on the multimodal feature $\mathbf{e}_i$ and the sampled environmental feature $\mathbf{\hat e}_i$, the veracity prediction objective is as follows:
\begin{equation}
    \label{eq2}
    \mathcal{L}_{VP} = \frac{1}{K} \sum \nolimits _{k=1}^K \frac{1}{|\mathcal{D}_k|} \sum \nolimits _{i=1}^{|\mathcal{D}_k|}
    \ell_{CE} \big( [ \mathbf{e}_i;\mathbf{\hat e}_i ] \mathbf{W}_C, y_i \big),
\end{equation}
where $\ell_{CE}(\cdot, \cdot)$ indicates a cross-entropy loss function, and $\mathbf{W}_C$ is a veracity classifier. Accordingly, the overall objective of \baby can be summarized as follows:
\begin{equation}
    \label{eq3}
    \mathcal{L} = \mathcal{L}_{VP} + \alpha \mathcal{L}_{CL} + \beta \mathcal{L}_{VG} + \gamma \mathcal{L}_{DM},
\end{equation}
where $\alpha$, $\beta$ and $\gamma$ are trade-off parameters to balance objectives. In the following sections, we present the details of the dynamically adapted MoE module and environmental dynamics model.

\subsection{Dynamically Adapted MoE}
Generally, the dynamically adapted MoE module aims to alleviate the knowledge forgetting issue by employing an MoE-based model, which consists of a Dirichlet process mixture-based method to dynamically expand experts and an exponential moving average optimized event-shared expert.

Specifically, we initialize a event-shared expert $\mathbf{E}_s$ and $M$ event-specific experts $\{\mathbf{E}_m\}_{m=1}^M$, and each expert $\mathbf{E}_m$ contains a discriminator part $\mathbf{E}_m^D$ and a variational generator part $\mathbf{E}_m^G$. The discriminator part is implemented as a low-rank linear LoRA model \citep{hu2022lora}, and the generator part is structured as a variational autoencoder \citep{kingma2014auto}, which includes an encoder that produces a hidden Gaussian distribution and a decoder that reconstructs the input provided to the encoder. 
Given a fused multimodal feature $\mathbf{z}_i$, we first decide whether it is necessary to expand a new event-specific expert $\mathbf{E}_{M+1}$.
Following Dirichlet process mixture in continual learning \citep{lee2020a,ye2021lifelong}, we define the responsibility $\rho_{i, m}$ of a new batched data features $\mathbf{z}_i$ as follows:
\begin{equation}
    \label{eq4}
    \renewcommand\arraystretch{1.3}
    \rho_{i, m} \varpropto \left\{
	\begin{array}{cc}
    	(\sum \rho_{<i, m}) \mathbb{P}(y_i | \mathbf{z}_i; \mathbf{E}_m^D) \mathbb{P}(\mathbf{z}_i; \mathbf{E}_m^G), & 1 \leq m < M,\\
    	\lambda \mathbb{P}(y_i | \mathbf{z}_i; \mathbf{E}_0^D) \mathbb{P}(\mathbf{z}_i; \mathbf{E}_0^G), & m = M + 1, \\
	\end{array}
	\right .
\end{equation}
where $\lambda$ is a hyper-parameter to control the model's sensitivity to new data, and $\mathbf{E}_0 = \{\mathbf{E}_0^D, \mathbf{E}_0^G\}$ represents the initialized weight of a new expert, we specify it as the weight of event-shared expert $\mathbf{E}_s = \{\mathbf{E}_s^D, \mathbf{E}_s^G\}$.
According to the responsibility, if $\arg \max_{m} \rho_{i, m} = M + 1$, we add a new expert, whose parameters are initialized by the shared expert $\mathbf{E}_s$.
In our specific implementation, we take the negative logarithm of this responsibility as follows:
\begin{equation}
    \label{eq5}
    \renewcommand{\arraystretch}{1.3}
    \begin{aligned}
        - \log & \rho_{i, m} \propto 
        \begin{cases}
            \begin{aligned}
                - \log \textstyle \sum \rho_{<i, m} - & \log \mathbb{P}(y_i | \mathbf{z}_i; \mathbf{E}_m^D) \\
                & - \log \mathbb{P}(\mathbf{z}_i; \mathbf{E}_m^G) ,
            \end{aligned} 
            & 1 \leq m < M, \\
            \begin{aligned}
                - \log \lambda - \log \mathbb{P}(y_i | & \mathbf{z}_i; \mathbf{E}_s^D) \\
                - & \log \mathbb{P}(\mathbf{z}_i; \mathbf{E}_s^G) ,
            \end{aligned}
            & m = M + 1,
        \end{cases} \\
        = & 
        \begin{cases}
            \begin{aligned}
                - \log \textstyle \sum \rho_{<i, m} + \ell_{CE} \big( [\mathbf{z}_i \mathbf{E}_m^D; \mathbf{\hat e}] \mathbf{W}_C, y_i\big) \\ 
                + \mathcal{L}_{VG},
            \end{aligned} 
            & 1 \leq m < M, \\
                - \log \lambda + \ell_{CE} \big( [\mathbf{z}_i \mathbf{E}_s^D; \mathbf{\hat e}] \mathbf{W}_C,  y_i \big) + \mathcal{L}_{VG},
            & m = M + 1.
        \end{cases}
    \end{aligned}
\end{equation}
where the term $- \log \mathbb{P}(y_i | \mathbf{z}_i; \mathbf{E}_m^D)$ is specified as a cross-entropy loss between the prediction of the expert $\mathbf{E}_m^D$ and the ground-truth veracity label $y_i$. And we specify the term $- \log \mathbb{P}(\mathbf{z}_i; \mathbf{E}_m^G)$ as a variational generation loss $\mathcal{L}_{VG}$, which is implemented by a reconstruction objective $\|\mathbf{z}_i \mathbf{E}_m^G - \mathbf{z}_i\|_2^2$. Accordingly, if the responsibility satisfies:
\begin{equation}
    \label{eq6}
    \arg \min_{m} - \log \rho_{i, m} = M + 1,
\end{equation}
a new expert will be created. Upon the experts, we obtain a forgetting resistant feature $\mathbf{e}_i$ as follows:
\begin{equation}
    \label{eq7}
    \mathbf{e}_i = r_i \mathbf{z}_i \mathbf{E}_{-1}^D + (1 - r_i) \mathbf{z}_i \mathbf{E}_s^D, \quad r_i = \mathbf{z}_i \mathbf{W}_R,
\end{equation}
where $\mathbf{W}_R$ is a routing network to assign an adaptive weight, and $\mathbf{E}_{-1}^D$ denotes the discriminator part of the latest event-specific expert. Additionally, since the event-shared expert $\mathbf{E}_s$ preserves the shared knowledge across all data, we employ an exponential moving average to update the event-shared expert to reduce the interference of new data with the old knowledge stored within it, as follows:
\begin{equation}
    \label{eq8}
    \mathbf{\tilde E}_s^D(\tau) = \epsilon \mathbf{\tilde E}_s^D(\tau - 1) + (1 - \epsilon) \mathbf{E}_s^D(\tau),
\end{equation}
where $\mathbf{\tilde E}_s^D(\tau)$ indicates the updated event-shared expert in the $\tau$-th updating step, and $\epsilon$ is a smoothing hyper-parameter, which is empirically fixed to 0.99 in our experiments.

\subsection{Environmental Dynamics Model}

We observe that when continually learning an MMD model, the social environment is consistently evolving over time. To predict the dynamics, we train a dynamics model to predict the environmental distribution of future samples, and sample an environmental feature $\mathbf{\hat e}_i$ for veracity prediction.

Specifically, given the multimodal feature $\mathbf{e}_i$ from the dynamically adapted MoE module, we map the samples with their labels $y=1$ to a shared environmental space and compute their mean and covariance matrix, thereby modeling the Gaussian distributions 
$\mathbb{P}(\mathbf{\hat e}|y=1) = \mathcal{N} \big( \boldsymbol{\mu}, \boldsymbol{\sigma}^2 \big)$ as follows:
\begin{equation}
    \label{eq9}
        \boldsymbol{\mu} = \frac{\sum \mathbf{e}_{y=1} \mathbf{W}_F}{|\mathbf{e}_{y=1}|} , \
        \boldsymbol{\sigma} = \frac{\sum (\mathbf{e}_{y=1}\mathbf{W}_F - \boldsymbol{\mu})(\mathbf{e}_{y=1}\mathbf{W}_F - \boldsymbol{\mu})^\top}{|\mathbf{e}_{y=1}|} ,
\end{equation}
where $\mathbf{W}_F$ is a learnable matrix to map multimodal features to the shared environmental space. 
Based on the environmental distribution specified by the Gaussian distribution, we supervise a dynamics model to predict the environmental distribution of future samples. Specifically, we take inspiration from neural ordinary differential equations \citep{chen2018neural,zang2020neural} to learn the following two time-dependent differentials, which represent the time derivative of the mean $\boldsymbol{\mu}$ and covariance $\boldsymbol{\sigma}$ of the Gaussian distribution as follows:
\begin{equation}
    \label{eq11}
    \frac{\mathrm{d} \boldsymbol{\mu}}{\mathrm{d} t} = \mathcal{G}_{\boldsymbol{\phi}_{\boldsymbol{\mu}}} (\boldsymbol{\mu}, t), \quad
    \frac{\mathrm{d} \boldsymbol{\sigma}}{\mathrm{d} t} = \mathcal{G}_{\boldsymbol{\phi}_{\boldsymbol{\sigma}}} (\boldsymbol{\sigma}, t).
\end{equation}
Upon this dynamics equation, we can predict the mean and covariance for any continuous time point $\tau$, thereby obtaining the environmental distribution $\mathbb{P}(\mathbf{\hat e}_\tau|y=1) = \mathcal{N} \big( \boldsymbol{\hat \mu}_\tau, \boldsymbol{\hat \sigma}_\tau^2 \big)$ at this time point as follows:
\begin{equation}
    \label{eq12}
    \boldsymbol{\hat \mu}_\tau = \boldsymbol{\mu}_0 + \int_0^\tau \mathcal{G}_{\boldsymbol{\phi}_{\boldsymbol{\mu}}} (\boldsymbol{\mu}, t) \mathrm{d} t, \quad
    \boldsymbol{\hat \sigma}_\tau = \boldsymbol{\sigma}_0 + \int_0^\tau \mathcal{G}_{\boldsymbol{\phi}_{\boldsymbol{\sigma}}} (\boldsymbol{\sigma}, t) \mathrm{d} t,
\end{equation}
where $\boldsymbol{\mu}_0$ and $\boldsymbol{\sigma}_0$ represent the initial state of $\boldsymbol{\mu}$ and $\boldsymbol{\sigma}$, and we specify them as the mean and covariance of the data in $\mathcal{D}_{k=1}$ following Eq.~\eqref{eq9}.
Then, to supervise the dynamics model, we assign a temporal label for each sample and reformulate the MMD dataset as  $\mathcal{D}_k = \{\mathcal{X}_i, y_i, t_i\}_{i=1}^{|\mathcal{D}_k|}, k \in \{1, \cdots, K\}$, where $t_i = k$ is the assigned temporal label. Following Eq.~\eqref{eq9}, we compute the mean $\boldsymbol{\mu}_\tau$ and covariance $\boldsymbol{\sigma}_\tau$ of each batched data whose $t_i = \tau$ as the ground-truth label, and formulate the following objective:
\begin{equation}
    \label{eq13}
    \small
    \begin{aligned}
        \mathcal{L}_{DM} = & \frac{1}{K} \int_0^K \| \boldsymbol{\hat \mu}_\tau - \boldsymbol{\mu}_\tau \|_2^2 + \| \boldsymbol{\hat \sigma}_\tau - \boldsymbol{\sigma}_\tau \|_2^2 \mathrm{d} t, \\
        \textbf{s.t.} \quad & \mathbf{H}_{\boldsymbol{\mu}_0} = \boldsymbol{\mu}_0 \mathbf{W}_E, \
            \mathbf{H}_{\boldsymbol{\sigma}_0} = \boldsymbol{\sigma}_0 \mathbf{W}_E, \quad \boldsymbol{\hat \mu}_\tau = \mathbf{H}_{\boldsymbol{\mu}_\tau} \mathbf{W}_D, \
            \boldsymbol{\hat \sigma}_\tau = \mathbf{H}_{\boldsymbol{\sigma}_\tau} \mathbf{W}_D,\\
        \quad \mathbf{H}_{\boldsymbol{\mu}_\tau} & = \mathbf{H}_{\boldsymbol{\mu}_0} + \int_0^\tau \mathcal{G}_{\boldsymbol{\phi}_{\boldsymbol{\mu}}}(\mathbf{H}_{\boldsymbol{\mu}}, t) \mathrm{d} t, 
        \ \mathbf{H}_{\boldsymbol{\sigma}_\tau} = \mathbf{H}_{\boldsymbol{\sigma}_0} + \int_0^\tau \mathcal{G}_{\boldsymbol{\phi}_{\boldsymbol{\sigma}}}(\mathbf{H}_{\boldsymbol{\sigma}}, t) \mathrm{d} t,\\
    \end{aligned}
\end{equation}
where $\mathbf{W}_E$ and $\mathbf{W}_D$ denote an encoder and decoder to map $\boldsymbol{\mu}_0$ and $\boldsymbol{\sigma}_0$ into a shared hidden space and decode them back, respectively. To compute the integral in Eq.~\eqref{eq13}, we use the \textit{dopri5} algorithm \citep{wanner1996solving} as the ordinary differential equation solver as follows:
\begin{equation}
    \label{eq14}
    \begin{aligned}
        \int_0^\tau & \mathcal{G}_{\boldsymbol{\phi}_{\boldsymbol{\mu}}}(\mathbf{H}_{\boldsymbol{\mu}}, t) \mathrm{d} t = dopri(\mathbf{H}_{\boldsymbol{\mu}}, \tau, \boldsymbol{\phi}_{\boldsymbol{\mu}}), \\
        \int_0^\tau & \mathcal{G}_{\boldsymbol{\phi}_{\boldsymbol{\sigma}}}(\mathbf{H}_{\boldsymbol{\sigma}}, t) \mathrm{d} t = dopri(\mathbf{H}_{\boldsymbol{\sigma}}, \tau, \boldsymbol{\phi}_{\boldsymbol{\sigma}}).
    \end{aligned}
\end{equation}
Upon the predicted distribution $\mathbb{P}(\mathbf{\hat e}_\tau|y=1) = \mathcal{N} \big( \boldsymbol{\hat \mu}_\tau, \boldsymbol{\hat \sigma}_\tau^2 \big)$ for the $i$-th sample from the $\tau$-th time period, we directly sample the environmental feature $\mathbf{\hat e}_i \sim \mathbb{P}(\mathbf{\hat e}_\tau|y=1)$ and concatenate it with $\mathbf{e}_i$ for veracity prediction as Eq.~\eqref{eq2}. The overall pseudo algorithm of \baby is depicted in the Supplementary Material.

\section{Experiments}

In this section, we conduct the comparative experiments to evaluate the performance of \baby.

\subsection{Experimental Settings}

\textbf{Datasets.}
We conduct experiments on three MMD datasets \textbf{\textit{GossipCop}} \citep{shu2020fakenewsnet}, \textbf{\textit{Weibo}} \citep{jin2017multimodal}, and \textbf{\textit{Twitter}} \citep{boididou2018detection}. 
To split them into continual MMD datasets, we follow EANN \citep{wang2018eann} to use the single-pass clustering \citep{jin2014news} to split three datasets into four subsets, respectively. Specifically, \textit{GossipCop} contains 12,840 English entertainment news articles, approximately 20\% of which are identified as fake news. \textit{Weibo} and \textit{Twitter} are derived from Chinese and English social media platforms, respectively. \textit{Weibo} comprises 9,528 image-text pairs, and \textit{Twitter} exhibits a significant imbalance between its text and image modalities, with numerous articles using overlapping images, making it particularly challenging.
For clarity, the detailed statistics of these datasets are shown in Table~\ref{datasetsta}.

\begin{table*}[t]
\centering
\renewcommand\arraystretch{0.95}
  \caption{Experimental results of \baby on three continual MMD datasets. The bold results represent the best performance and they are statistically significant than the baseline models.}
  \label{result}
  \small
  \setlength{\tabcolsep}{5pt}{
  \begin{tabular}{m{1.91cm}m{0.94cm}<{\centering}m{0.94cm}<{\centering}m{0.94cm}<{\centering}m{0.94cm}<{\centering}m{0.94cm}<{\centering}m{0.94cm}<{\centering}m{0.94cm}<{\centering}m{0.94cm}<{\centering}m{0.94cm}<{\centering}m{0.94cm}<{\centering}m{0.94cm}<{\centering}m{0.94cm}<{\centering}}
    \toprule
    \multirow{2}{*}{\quad \quad Model} & \multicolumn{4}{c}{\textbf{Dataset}: \textit{GossipCop} \citep{shu2020fakenewsnet}} & \multicolumn{4}{c}{\textbf{Dataset}: \textit{Weibo} \citep{jin2017multimodal}} & \multicolumn{4}{c}{\textbf{Dataset}: \textit{Twitter} \citep{boididou2018detection}} \\
    
    \cmidrule(r){2-5} \cmidrule(r){6-9} \cmidrule(r){10-13} 
    & Accuracy & F1 & F1$_{\text{real}}$ & F1$_{\text{fake}}$ & Accuracy & F1 & F1$_{\text{real}}$ & F1$_{\text{fake}}$ & Accuracy & F1 & F1$_{\text{real}}$ & F1$_{\text{fake}}$ \\
    \hline
    \textbf{Base} \citep{he2016deep,devlin2019bert} & 84.52{\color{grayv} \footnotesize $\pm$0.7} & 73.32{\color{grayv} \footnotesize $\pm$1.3} & 90.61{\color{grayv} \footnotesize $\pm$0.5} & 57.02{\color{grayv} \footnotesize $\pm$1.8} & 
    87.98{\color{grayv} \footnotesize $\pm$0.7} & 87.94{\color{grayv} \footnotesize $\pm$0.7} & 87.27{\color{grayv} \footnotesize $\pm$0.8} & 88.60{\color{grayv} \footnotesize $\pm$0.7} & 
    62.20{\color{grayv} \footnotesize $\pm$1.8} & 62.10{\color{grayv} \footnotesize $\pm$1.8} & 60.98{\color{grayv} \footnotesize $\pm$1.9} & 63.22{\color{grayv} \footnotesize $\pm$1.8} \\
    \ + EWC \citep{kirkpatrick2016overcoming,han2021continual} & 84.91{\color{grayv} \footnotesize $\pm$0.6} & 74.95{\color{grayv} \footnotesize $\pm$2.2} & 90.69{\color{grayv} \footnotesize $\pm$0.5} & 58.81{\color{grayv} \footnotesize $\pm$1.6} & 
    88.46{\color{grayv} \footnotesize $\pm$0.5} & 88.41{\color{grayv} \footnotesize $\pm$0.5} & 87.64{\color{grayv} \footnotesize $\pm$0.6} & 89.19{\color{grayv} \footnotesize $\pm$0.5} & 
    64.13{\color{grayv} \footnotesize $\pm$1.1} & 63.73{\color{grayv} \footnotesize $\pm$1.4} & 60.68{\color{grayv} \footnotesize $\pm$1.8} & 66.78{\color{grayv} \footnotesize $\pm$2.1} \\
    \ + Replay \citep{lee2021dynamically} & 84.90{\color{grayv} \footnotesize $\pm$1.2} & 74.81{\color{grayv} \footnotesize $\pm$2.2} & 90.68{\color{grayv} \footnotesize $\pm$0.7} & 59.00{\color{grayv} \footnotesize $\pm$1.7} & 
    88.53{\color{grayv} \footnotesize $\pm$0.8} & 88.47{\color{grayv} \footnotesize $\pm$0.8} & 87.65{\color{grayv} \footnotesize $\pm$0.8} & 89.30{\color{grayv} \footnotesize $\pm$0.9} & 
    63.90{\color{grayv} \footnotesize $\pm$1.8} & 63.82{\color{grayv} \footnotesize $\pm$1.7} & 62.58{\color{grayv} \footnotesize $\pm$1.2} & 65.05{\color{grayv} \footnotesize $\pm$1.4} \\
    \ + LoRAMoE \citep{dou2024loramoe} & 84.89{\color{grayv} \footnotesize $\pm$1.2} & 74.91{\color{grayv} \footnotesize $\pm$0.8} & 90.67{\color{grayv} \footnotesize $\pm$0.9} & 59.18{\color{grayv} \footnotesize $\pm$1.4} & 
    88.60{\color{grayv} \footnotesize $\pm$0.9} & 88.55{\color{grayv} \footnotesize $\pm$0.9} & 87.75{\color{grayv} \footnotesize $\pm$1.0} & 89.34{\color{grayv} \footnotesize $\pm$0.9} & 
    64.64{\color{grayv} \footnotesize $\pm$1.3} & 64.56{\color{grayv} \footnotesize $\pm$1.3} & 62.91{\color{grayv} \footnotesize $\pm$1.8} & 66.21{\color{grayv} \footnotesize $\pm$1.9} \\
    
    \rowcolor{lightgrayv} \ + \textbf{\baby} & \textbf{86.21}{\color{grayv} \footnotesize $\pm$0.8} & \textbf{76.13}{\color{grayv} \footnotesize $\pm$0.5} & \textbf{91.64}{\color{grayv} \footnotesize $\pm$0.6} & \textbf{60.63}{\color{grayv} \footnotesize $\pm$1.0} & 
    \textbf{90.07}{\color{grayv} \footnotesize $\pm$0.6} & \textbf{90.05}{\color{grayv} \footnotesize $\pm$0.6} & \textbf{89.67}{\color{grayv} \footnotesize $\pm$0.6} & \textbf{90.50}{\color{grayv} \footnotesize $\pm$0.6} & 
    \textbf{67.66}{\color{grayv} \footnotesize $\pm$1.7} & \textbf{67.44}{\color{grayv} \footnotesize $\pm$1.5} & \textbf{65.64}{\color{grayv} \footnotesize $\pm$1.3} & \textbf{69.24}{\color{grayv} \footnotesize $\pm$2.0} \\
    \hline
    
    \textbf{SAFE} \citep{zhou2020safe} & 84.31{\color{grayv} \footnotesize $\pm$1.4} & 73.93{\color{grayv} \footnotesize $\pm$0.6} & 90.38{\color{grayv} \footnotesize $\pm$1.1} & 57.47{\color{grayv} \footnotesize $\pm$1.2} & 
    87.30{\color{grayv} \footnotesize $\pm$1.2} & 87.16{\color{grayv} \footnotesize $\pm$1.2} & 85.82{\color{grayv} \footnotesize $\pm$1.1} & 88.50{\color{grayv} \footnotesize $\pm$1.1} & 
    62.46{\color{grayv} \footnotesize $\pm$1.8} & 61.69{\color{grayv} \footnotesize $\pm$2.4} & 58.49{\color{grayv} \footnotesize $\pm$1.7} & 66.89{\color{grayv} \footnotesize $\pm$2.1} \\
    \ + EWC \citep{kirkpatrick2016overcoming,han2021continual} & 84.63{\color{grayv} \footnotesize $\pm$0.7} & 74.48{\color{grayv} \footnotesize $\pm$0.7} & 90.85{\color{grayv} \footnotesize $\pm$1.5} & 59.12{\color{grayv} \footnotesize $\pm$1.6} & 
    87.90{\color{grayv} \footnotesize $\pm$0.9} & 87.88{\color{grayv} \footnotesize $\pm$0.9} & 87.43{\color{grayv} \footnotesize $\pm$0.9} & 88.32{\color{grayv} \footnotesize $\pm$0.8} & 
    63.25{\color{grayv} \footnotesize $\pm$1.4} & 62.90{\color{grayv} \footnotesize $\pm$1.4} & 60.33{\color{grayv} \footnotesize $\pm$1.9} & 65.48{\color{grayv} \footnotesize $\pm$1.9} \\
    \ + Replay \citep{lee2021dynamically} & 84.81{\color{grayv} \footnotesize $\pm$0.7} & 74.68{\color{grayv} \footnotesize $\pm$1.1} & 90.54{\color{grayv} \footnotesize $\pm$0.6} & 58.76{\color{grayv} \footnotesize $\pm$2.2} & 
    88.67{\color{grayv} \footnotesize $\pm$1.0} & 88.60{\color{grayv} \footnotesize $\pm$1.0} & 87.72{\color{grayv} \footnotesize $\pm$1.0} & 89.49{\color{grayv} \footnotesize $\pm$1.0} & 
    64.01{\color{grayv} \footnotesize $\pm$0.6} & 62.73{\color{grayv} \footnotesize $\pm$1.2} & 59.73{\color{grayv} \footnotesize $\pm$2.0} & 67.73{\color{grayv} \footnotesize $\pm$1.9} \\
    \ + LoRAMoE \citep{dou2024loramoe} & 84.66{\color{grayv} \footnotesize $\pm$0.8} & 75.06{\color{grayv} \footnotesize $\pm$0.8} & 90.54{\color{grayv} \footnotesize $\pm$0.6} & 58.59{\color{grayv} \footnotesize $\pm$1.1} & 
    88.67{\color{grayv} \footnotesize $\pm$0.8} & 88.64{\color{grayv} \footnotesize $\pm$0.9} & 88.04{\color{grayv} \footnotesize $\pm$0.7} & 89.23{\color{grayv} \footnotesize $\pm$0.4} & 
    64.73{\color{grayv} \footnotesize $\pm$1.6} & 64.19{\color{grayv} \footnotesize $\pm$1.4} & 59.80{\color{grayv} \footnotesize $\pm$1.3} & 67.58{\color{grayv} \footnotesize $\pm$1.6} \\
    
    \rowcolor{lightgrayv} \ + \textbf{\baby} & \textbf{86.72}{\color{grayv} \footnotesize $\pm$0.4} & \textbf{76.48}{\color{grayv} \footnotesize $\pm$1.2} & \textbf{91.99}{\color{grayv} \footnotesize $\pm$0.2} & \textbf{60.98}{\color{grayv} \footnotesize $\pm$1.3} &
    \textbf{90.46}{\color{grayv} \footnotesize $\pm$0.2} & \textbf{90.45}{\color{grayv} \footnotesize $\pm$0.2} & \textbf{90.19}{\color{grayv} \footnotesize $\pm$0.3} & \textbf{90.72}{\color{grayv} \footnotesize $\pm$0.1} & 
    \textbf{68.08}{\color{grayv} \footnotesize $\pm$1.8} & \textbf{67.36}{\color{grayv} \footnotesize $\pm$1.5} & \textbf{63.65}{\color{grayv} \footnotesize $\pm$1.3} & \textbf{71.07}{\color{grayv} \footnotesize $\pm$1.9} \\
    \hline

    \textbf{MCAN} \citep{wu2021multimodal} & 84.54{\color{grayv} \footnotesize $\pm$1.9} & 73.31{\color{grayv} \footnotesize $\pm$0.8} & 90.87{\color{grayv} \footnotesize $\pm$0.4} & 60.19{\color{grayv} \footnotesize $\pm$1.3} & 
    87.71{\color{grayv} \footnotesize $\pm$1.0} & 87.61{\color{grayv} \footnotesize $\pm$1.0} & 86.47{\color{grayv} \footnotesize $\pm$1.3} & 88.75{\color{grayv} \footnotesize $\pm$0.8} &
    62.98{\color{grayv} \footnotesize $\pm$2.0} & 60.58{\color{grayv} \footnotesize $\pm$1.2} & 58.27{\color{grayv} \footnotesize $\pm$1.4} & 68.90{\color{grayv} \footnotesize $\pm$2.1} \\
    \ + EWC \citep{kirkpatrick2016overcoming,han2021continual} & 85.21{\color{grayv} \footnotesize $\pm$0.8} & 73.44{\color{grayv} \footnotesize $\pm$1.2} & 90.84{\color{grayv} \footnotesize $\pm$0.9} & 57.40{\color{grayv} \footnotesize $\pm$2.2} & 
    88.40{\color{grayv} \footnotesize $\pm$0.6} & 88.38{\color{grayv} \footnotesize $\pm$0.6} & 87.98{\color{grayv} \footnotesize $\pm$0.6} & 88.79{\color{grayv} \footnotesize $\pm$0.6} & 
    64.37{\color{grayv} \footnotesize $\pm$1.0} & 64.26{\color{grayv} \footnotesize $\pm$0.6} & 60.59{\color{grayv} \footnotesize $\pm$1.3} & 66.98{\color{grayv} \footnotesize $\pm$2.1} \\
    \ + Replay \citep{lee2021dynamically} & 85.19{\color{grayv} \footnotesize $\pm$0.7} & 73.25{\color{grayv} \footnotesize $\pm$1.0} & 90.72{\color{grayv} \footnotesize $\pm$1.0} & 58.78{\color{grayv} \footnotesize $\pm$2.1} & 
    88.53{\color{grayv} \footnotesize $\pm$0.7} & 88.51{\color{grayv} \footnotesize $\pm$0.7} & 87.97{\color{grayv} \footnotesize $\pm$0.9} & 89.05{\color{grayv} \footnotesize $\pm$0.7} & 
    64.46{\color{grayv} \footnotesize $\pm$1.0} & 64.01{\color{grayv} \footnotesize $\pm$1.3} & 60.02{\color{grayv} \footnotesize $\pm$1.4} & 67.01{\color{grayv} \footnotesize $\pm$1.3} \\
    \ + LoRAMoE \citep{dou2024loramoe} & 84.88{\color{grayv} \footnotesize $\pm$0.8} & 74.38{\color{grayv} \footnotesize $\pm$0.6} & 90.89{\color{grayv} \footnotesize $\pm$0.5} & 59.30{\color{grayv} \footnotesize $\pm$0.8} & 
    87.99{\color{grayv} \footnotesize $\pm$0.7} & 87.94{\color{grayv} \footnotesize $\pm$0.7} & 87.17{\color{grayv} \footnotesize $\pm$0.8} & 88.70{\color{grayv} \footnotesize $\pm$0.6} & 
    64.64{\color{grayv} \footnotesize $\pm$1.4} & 64.63{\color{grayv} \footnotesize $\pm$1.3} & 60.62{\color{grayv} \footnotesize $\pm$1.3} & 67.68{\color{grayv} \footnotesize $\pm$2.0} \\
    
    \rowcolor{lightgrayv} \ + \textbf{\baby} & \textbf{86.37}{\color{grayv} \footnotesize $\pm$0.7} & \textbf{76.33}{\color{grayv} \footnotesize $\pm$0.8} & \textbf{91.74}{\color{grayv} \footnotesize $\pm$0.5} & \textbf{60.91}{\color{grayv} \footnotesize $\pm$1.3} &
    \textbf{90.10}{\color{grayv} \footnotesize $\pm$0.8} & \textbf{90.07}{\color{grayv} \footnotesize $\pm$0.8} & \textbf{89.58}{\color{grayv} \footnotesize $\pm$0.9} & \textbf{90.58}{\color{grayv} \footnotesize $\pm$0.7} & 
    \textbf{67.43}{\color{grayv} \footnotesize $\pm$0.9} & \textbf{66.99}{\color{grayv} \footnotesize $\pm$0.9} & \textbf{63.81}{\color{grayv} \footnotesize $\pm$1.3} & \textbf{71.18}{\color{grayv} \footnotesize $\pm$1.3} \\
    \hline

    \textbf{CAFE} \citep{chen2022cross} & 84.03{\color{grayv} \footnotesize $\pm$1.3} & 74.46{\color{grayv} \footnotesize $\pm$0.8} & 90.09{\color{grayv} \footnotesize $\pm$1.0} & 59.21{\color{grayv} \footnotesize $\pm$1.5} & 
    87.81{\color{grayv} \footnotesize $\pm$1.0} & 87.75{\color{grayv} \footnotesize $\pm$1.0} & 86.94{\color{grayv} \footnotesize $\pm$1.2} & 88.55{\color{grayv} \footnotesize $\pm$0.8} & 
    61.77{\color{grayv} \footnotesize $\pm$1.5} & 60.96{\color{grayv} \footnotesize $\pm$1.3} & 58.62{\color{grayv} \footnotesize $\pm$1.5} & 66.31{\color{grayv} \footnotesize $\pm$2.0} \\
    \ + EWC \citep{kirkpatrick2016overcoming,han2021continual} & 84.63{\color{grayv} \footnotesize $\pm$1.2} & 75.30{\color{grayv} \footnotesize $\pm$0.8} & 90.48{\color{grayv} \footnotesize $\pm$0.9} & 60.13{\color{grayv} \footnotesize $\pm$0.9} & 
    88.50{\color{grayv} \footnotesize $\pm$1.0} & 88.42{\color{grayv} \footnotesize $\pm$1.1} & 87.43{\color{grayv} \footnotesize $\pm$1.3} & 89.40{\color{grayv} \footnotesize $\pm$0.8} & 
    63.83{\color{grayv} \footnotesize $\pm$1.2} & 63.19{\color{grayv} \footnotesize $\pm$1.7} & 59.30{\color{grayv} \footnotesize $\pm$1.5} & 67.07{\color{grayv} \footnotesize $\pm$1.8} \\
    \ + Replay \citep{lee2021dynamically} & 84.59{\color{grayv} \footnotesize $\pm$1.0} & 74.65{\color{grayv} \footnotesize $\pm$1.4} & 90.54{\color{grayv} \footnotesize $\pm$0.9} & 59.76{\color{grayv} \footnotesize $\pm$1.8} & 
    88.03{\color{grayv} \footnotesize $\pm$0.5} & 88.01{\color{grayv} \footnotesize $\pm$0.5} & 87.55{\color{grayv} \footnotesize $\pm$0.6} & 88.47{\color{grayv} \footnotesize $\pm$0.5} & 
    65.06{\color{grayv} \footnotesize $\pm$1.2} & 64.75{\color{grayv} \footnotesize $\pm$1.0} & 61.67{\color{grayv} \footnotesize $\pm$1.7} & 67.84{\color{grayv} \footnotesize $\pm$1.9} \\
    \ + LoRAMoE \citep{dou2024loramoe} & 84.87{\color{grayv} \footnotesize $\pm$0.7} & 74.00{\color{grayv} \footnotesize $\pm$1.6} & 90.60{\color{grayv} \footnotesize $\pm$0.3} & 59.67{\color{grayv} \footnotesize $\pm$2.5} & 
    88.60{\color{grayv} \footnotesize $\pm$0.6} & 88.52{\color{grayv} \footnotesize $\pm$0.6} & 87.53{\color{grayv} \footnotesize $\pm$0.8} & 89.50{\color{grayv} \footnotesize $\pm$0.9} & 
    64.53{\color{grayv} \footnotesize $\pm$1.8} & 62.08{\color{grayv} \footnotesize $\pm$1.4} & 60.26{\color{grayv} \footnotesize $\pm$1.5} & 67.89{\color{grayv} \footnotesize $\pm$1.5} \\
    
    \rowcolor{lightgrayv} \ + \textbf{\baby} & \textbf{86.67}{\color{grayv} \footnotesize $\pm$0.2} & \textbf{76.50}{\color{grayv} \footnotesize $\pm$0.4} & \textbf{91.96}{\color{grayv} \footnotesize $\pm$0.1} & \textbf{61.05}{\color{grayv} \footnotesize $\pm$0.8} & 
    \textbf{90.00}{\color{grayv} \footnotesize $\pm$0.4} & \textbf{89.98}{\color{grayv} \footnotesize $\pm$0.4} & \textbf{89.53}{\color{grayv} \footnotesize $\pm$0.5} & \textbf{90.43}{\color{grayv} \footnotesize $\pm$0.4} & 
    \textbf{67.99}{\color{grayv} \footnotesize $\pm$1.3} & \textbf{67.21}{\color{grayv} \footnotesize $\pm$1.5} & \textbf{62.90}{\color{grayv} \footnotesize $\pm$1.2} & \textbf{71.52}{\color{grayv} \footnotesize $\pm$1.6} \\
    \hline

    \textbf{BMR} \citep{ying2023bootstrapping} & 83.92{\color{grayv} \footnotesize $\pm$1.6} & 73.64{\color{grayv} \footnotesize $\pm$1.0} & 90.60{\color{grayv} \footnotesize $\pm$1.2} & 58.10{\color{grayv} \footnotesize $\pm$2.0} & 
    87.54{\color{grayv} \footnotesize $\pm$0.8} & 87.50{\color{grayv} \footnotesize $\pm$0.8} & 86.84{\color{grayv} \footnotesize $\pm$0.9} & 88.16{\color{grayv} \footnotesize $\pm$0.8} & 
    62.22{\color{grayv} \footnotesize $\pm$1.2} & 61.31{\color{grayv} \footnotesize $\pm$1.2} & 58.39{\color{grayv} \footnotesize $\pm$1.8} & 67.24{\color{grayv} \footnotesize $\pm$1.9} \\
    \ + EWC \citep{kirkpatrick2016overcoming,han2021continual} & 84.45{\color{grayv} \footnotesize $\pm$1.0} & 74.58{\color{grayv} \footnotesize $\pm$0.9} & 90.22{\color{grayv} \footnotesize $\pm$0.8} & 59.63{\color{grayv} \footnotesize $\pm$2.2} & 
    88.40{\color{grayv} \footnotesize $\pm$0.8} & 88.34{\color{grayv} \footnotesize $\pm$0.8} & 87.52{\color{grayv} \footnotesize $\pm$0.6} & 89.16{\color{grayv} \footnotesize $\pm$1.0} & 
    63.49{\color{grayv} \footnotesize $\pm$1.5} & 63.23{\color{grayv} \footnotesize $\pm$1.4} & 60.12{\color{grayv} \footnotesize $\pm$1.5} & 66.34{\color{grayv} \footnotesize $\pm$2.0} \\
    \ + Replay \citep{lee2021dynamically} & 84.63{\color{grayv} \footnotesize $\pm$1.4} & 74.28{\color{grayv} \footnotesize $\pm$1.8} & 90.12{\color{grayv} \footnotesize $\pm$1.1} & 59.38{\color{grayv} \footnotesize $\pm$2.5} & 
    87.92{\color{grayv} \footnotesize $\pm$1.1} & 87.89{\color{grayv} \footnotesize $\pm$1.3} & 87.39{\color{grayv} \footnotesize $\pm$1.4} & 88.39{\color{grayv} \footnotesize $\pm$0.7} & 
    64.13{\color{grayv} \footnotesize $\pm$1.9} & 62.64{\color{grayv} \footnotesize $\pm$1.5} & 59.76{\color{grayv} \footnotesize $\pm$1.9} & 67.52{\color{grayv} \footnotesize $\pm$1.1} \\
    \ + LoRAMoE \citep{dou2024loramoe} & 84.81{\color{grayv} \footnotesize $\pm$0.9} & 74.67{\color{grayv} \footnotesize $\pm$0.8} & 90.46{\color{grayv} \footnotesize $\pm$0.8} & 58.87{\color{grayv} \footnotesize $\pm$1.8} & 
    88.26{\color{grayv} \footnotesize $\pm$1.5} & 88.20{\color{grayv} \footnotesize $\pm$1.5} & 87.37{\color{grayv} \footnotesize $\pm$1.5} & 89.03{\color{grayv} \footnotesize $\pm$1.5} & 
    64.10{\color{grayv} \footnotesize $\pm$1.7} & 63.58{\color{grayv} \footnotesize $\pm$1.6} & 59.21{\color{grayv} \footnotesize $\pm$1.4} & 67.45{\color{grayv} \footnotesize $\pm$2.2} \\
    
    \rowcolor{lightgrayv} \ + \textbf{\baby} & \textbf{86.13}{\color{grayv} \footnotesize $\pm$0.6} & \textbf{76.31}{\color{grayv} \footnotesize $\pm$0.7} & \textbf{91.56}{\color{grayv} \footnotesize $\pm$0.4} & \textbf{61.07}{\color{grayv} \footnotesize $\pm$1.2} & 
    \textbf{89.86}{\color{grayv} \footnotesize $\pm$0.8} & \textbf{89.84}{\color{grayv} \footnotesize $\pm$0.8} & \textbf{89.47}{\color{grayv} \footnotesize $\pm$1.1} & \textbf{90.20}{\color{grayv} \footnotesize $\pm$0.5} & 
    \textbf{67.36}{\color{grayv} \footnotesize $\pm$1.3} & \textbf{67.23}{\color{grayv} \footnotesize $\pm$1.3} & \textbf{66.08}{\color{grayv} \footnotesize $\pm$1.4} & \textbf{69.38}{\color{grayv} \footnotesize $\pm$1.5} \\
    \hline

    \textbf{GAMED} \citep{shen2025gamed} & 84.28{\color{grayv} \footnotesize $\pm$1.2} & 73.84{\color{grayv} \footnotesize $\pm$1.1} & 90.78{\color{grayv} \footnotesize $\pm$0.9} & 59.61{\color{grayv} \footnotesize $\pm$2.1} & 
    87.29{\color{grayv} \footnotesize $\pm$0.9} & 87.24{\color{grayv} \footnotesize $\pm$0.9} & 87.48{\color{grayv} \footnotesize $\pm$1.0} & 87.98{\color{grayv} \footnotesize $\pm$0.9} & 
    61.90{\color{grayv} \footnotesize $\pm$0.6} & 60.88{\color{grayv} \footnotesize $\pm$1.4} & 59.07{\color{grayv} \footnotesize $\pm$1.7} & 65.71{\color{grayv} \footnotesize $\pm$1.9} \\
    \ + EWC \citep{kirkpatrick2016overcoming,han2021continual} & 84.45{\color{grayv} \footnotesize $\pm$1.6} & 74.65{\color{grayv} \footnotesize $\pm$1.0} & 90.41{\color{grayv} \footnotesize $\pm$1.2} & 59.29{\color{grayv} \footnotesize $\pm$1.6} & 
    88.49{\color{grayv} \footnotesize $\pm$0.9} & 88.47{\color{grayv} \footnotesize $\pm$0.9} & 88.04{\color{grayv} \footnotesize $\pm$1.1} & 88.91{\color{grayv} \footnotesize $\pm$0.8} & 
    64.73{\color{grayv} \footnotesize $\pm$1.7} & 63.34{\color{grayv} \footnotesize $\pm$1.3} & 59.22{\color{grayv} \footnotesize $\pm$2.1} & 68.46{\color{grayv} \footnotesize $\pm$2.1} \\
    \ + Replay \citep{lee2021dynamically} & 84.68{\color{grayv} \footnotesize $\pm$2.0} & 74.09{\color{grayv} \footnotesize $\pm$0.5} & 90.19{\color{grayv} \footnotesize $\pm$1.6} & 59.27{\color{grayv} \footnotesize $\pm$1.9} & 
    87.91{\color{grayv} \footnotesize $\pm$0.8} & 87.89{\color{grayv} \footnotesize $\pm$0.8} & 87.47{\color{grayv} \footnotesize $\pm$0.8} & 88.30{\color{grayv} \footnotesize $\pm$0.7} & 
    64.20{\color{grayv} \footnotesize $\pm$1.9} & 63.55{\color{grayv} \footnotesize $\pm$1.7} & 59.81{\color{grayv} \footnotesize $\pm$1.0} & 67.29{\color{grayv} \footnotesize $\pm$1.5} \\
    \ + LoRAMoE \citep{dou2024loramoe} & 84.79{\color{grayv} \footnotesize $\pm$0.7} & 74.44{\color{grayv} \footnotesize $\pm$2.2} & 90.61{\color{grayv} \footnotesize $\pm$0.9} & 59.27{\color{grayv} \footnotesize $\pm$1.7} & 
    88.46{\color{grayv} \footnotesize $\pm$0.8} & 88.37{\color{grayv} \footnotesize $\pm$0.8} & 87.32{\color{grayv} \footnotesize $\pm$0.6} & 89.02{\color{grayv} \footnotesize $\pm$1.4} & 
    63.88{\color{grayv} \footnotesize $\pm$1.6} & 63.62{\color{grayv} \footnotesize $\pm$1.5} & 60.21{\color{grayv} \footnotesize $\pm$1.8} & 66.03{\color{grayv} \footnotesize $\pm$2.1} \\
    
    \rowcolor{lightgrayv} \ + \textbf{\baby} & \textbf{86.21}{\color{grayv} \footnotesize $\pm$1.0} & \textbf{76.17}{\color{grayv} \footnotesize $\pm$0.7} & \textbf{91.73}{\color{grayv} \footnotesize $\pm$0.5} & \textbf{60.62}{\color{grayv} \footnotesize $\pm$1.0} &
    \textbf{89.93}{\color{grayv} \footnotesize $\pm$0.7} & \textbf{89.90}{\color{grayv} \footnotesize $\pm$0.7} & \textbf{89.42}{\color{grayv} \footnotesize $\pm$0.9} & \textbf{90.39}{\color{grayv} \footnotesize $\pm$0.5} & 
    \textbf{67.21}{\color{grayv} \footnotesize $\pm$1.7} & \textbf{67.31}{\color{grayv} \footnotesize $\pm$1.3} & \textbf{62.98}{\color{grayv} \footnotesize $\pm$1.7} & \textbf{70.63}{\color{grayv} \footnotesize $\pm$1.5} \\
    
    \bottomrule
  \end{tabular} }
\end{table*}

\begin{table}[t]
\centering
\renewcommand\arraystretch{0.95}
\small
  \caption{Statistics of three prevalent continual MMD datasets.}
  \label{datasetsta}
  \setlength{\tabcolsep}{5pt}{
  \begin{tabular}{m{1.7cm}<{\centering}m{0.85cm}<{\centering}m{0.85cm}<{\centering}m{0.6cm}<{\centering}m{0.6cm}<{\centering}m{0.6cm}<{\centering}m{0.6cm}<{\centering}}
    \toprule
    \multirow{2}{*}{Dataset} & \multirow{2}{*}{Articles} & \multirow{2}{*}{Images} & \multicolumn{4}{c}{Training Subsets} \\
    \cmidrule(lr){4-7}
    &  &  & 1 & 2 & 3 & 4 \\
    \hline
    \textit{GossipCop} \citep{shu2020fakenewsnet} & 12,840 & 12,840 & 4,184 & 4,407 & 348 & 1,071 \\
    \textit{Weibo} \citep{jin2017multimodal} & 9,528 & 9,528 & 1,006 & 660 & 861 & 2,888 \\
    \textit{Twitter} \citep{boididou2018detection} & 13,924 & 514 & 3,176 & 862 & 3,736 & 5,983 \\
    \bottomrule
  \end{tabular} }
\end{table}

\vspace{2pt} \noindent
\textbf{Baselines methods.}
We select 6 MMD models as baselines to evaluate that our proposed method can enhance their performance in continuous data scenarios, including \textbf{Base} \citep{he2016deep,devlin2019bert}, \textbf{SAFE} \citep{zhou2020safe}, \textbf{MCAN} \citep{wu2021multimodal}, \textbf{CAFE} \citep{chen2022cross}, \textbf{BMR} \citep{ying2023bootstrapping}, and \textbf{GAMED} \citep{shen2025gamed}. 
Additionally, we compare three prevalent continual learning methods, including \textbf{EWC} \citep{kirkpatrick2016overcoming,han2021continual}, \textbf{Replay} \citep{robins1995catastrophic,lee2021dynamically}, and \textbf{LoRAMoE} \citep{dou2024loramoe}, which have been adapted to the continual MMD task. 

\vspace{2pt} \noindent
\textbf{Implementation details.}
For the MMD data, we resize the images to $224 \times 224$ and truncate or pad the text to a length of 128 tokens. We use the pre-trained BERT\footnote{\url{https://huggingface.co/google-bert/bert-base-uncased}.} \citep{devlin2019bert} model and a ResNet34 \citep{he2016deep} model to extract the semantic features of text and image modalities. For BERT, only the parameters of the last 3 Transformer layers are optimized. During training, we fine-tune the BERT parameters using the Adam optimizer with a learning rate of $3 \times 10^{-5}$, and the other modules in the model are optimized using the Adam optimizer with a learning rate of $1 \times 10^{-3}$. Additionally, we fix the batch size at 32 and employ an early stopping strategy, which means that if no better Macro F1 score is observed for 5 consecutive epochs on a subset, the training for that subset is stopped. For the ordinary differential equation solver in the dynamics model, we utilize an open-source toolkit\footnote{\url{https://github.com/rtqichen/torchdiffeq}.} \citep{chen2018neural}. As for the hyper-parameters in Eq.~\eqref{eq3}, we empirically fix $\alpha$, $\beta$, and $\gamma$ to 0.1, 0.1, and 1, respectively.

\subsection{Main Results}

We compare our method \baby with six baseline MMD models and three continual learning approaches across three continual MMD datasets. The experimental results are reported in Table~\ref{result}. All experiments are repeated five times using five different random seeds $\{1, 2, 3, 4, 5\}$, and their average values and standard deviations are reported in Table~\ref{result}.
Overall, our method \baby demonstrates consistently and significantly better performance in detecting misinformation compared to its baseline models and state-of-the-art continual learning approaches. For example, when applying \baby to the latest MMD baseline GAMED, the model’s detection accuracy improves from 84.28 to 86.21, and surpasses the state-of-the-art continual learning method LoRAMoE by approximately 1.42. This indicates that our method can enhance the model’s detection performance by mitigating catastrophic forgetting and predicting environmental distributions. 
Furthermore, turning to compare the results across three datasets, the improvements of our model over the baseline models can be roughly ranked as \textit{GossipCop} < \textit{Weibo} < \textit{Twitter}. This order is consistent with the trend in dataset scale, where the \textit{Twitter} dataset, being the smallest in scale, achieves the largest improvement. For example, on the GAMED baseline, the accuracy gains of our model on the three datasets are 1.93, 2.64, and 5.31, respectively. This phenomenon suggests that our method compensates for the lack of semantic information caused by limited training samples by incorporating additional dynamics information and leveraging an MoE system to capture deeper information. 
Meanwhile, the \textit{Twitter} dataset is naturally split into event-based subsets with significant differences between events. Therefore, the MoE system in \baby effectively isolates the training parameters across these subsets, preventing interference between them. This further highlights the adaptability and effectiveness of our method in handling continual data distributions.

\begin{figure*}[t]
  \centering
  \includegraphics[scale=0.202]{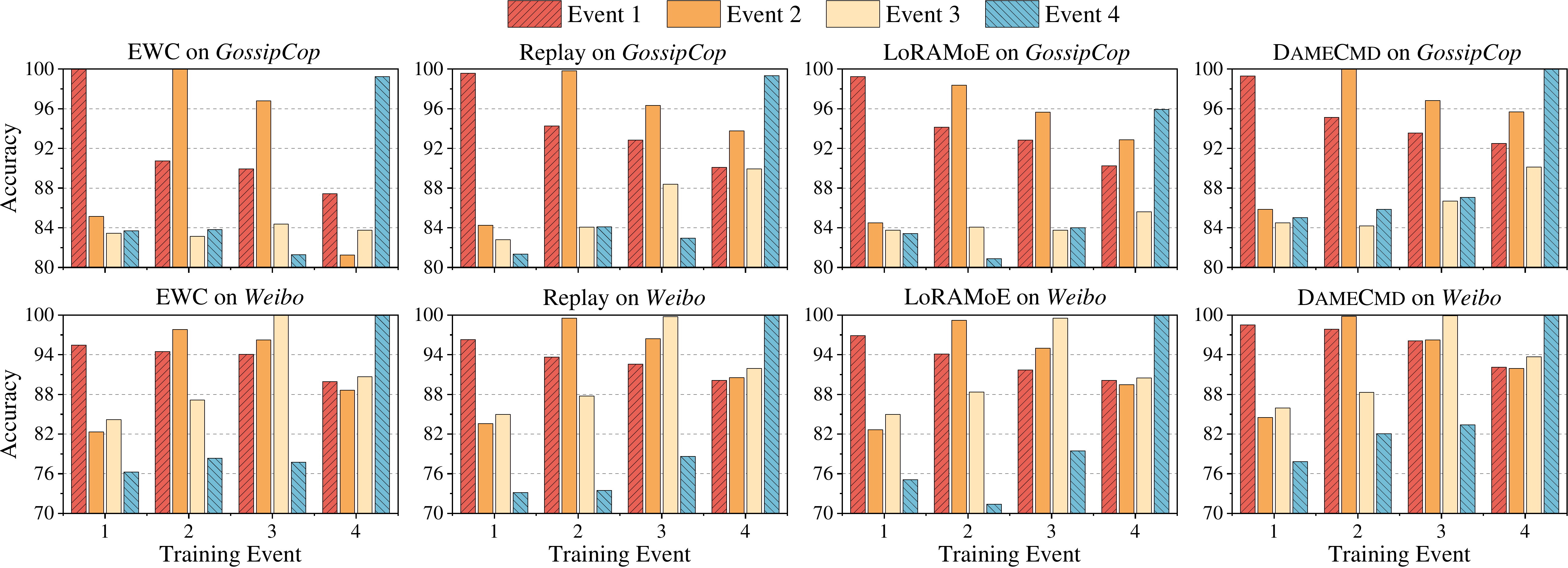}  
  \caption{The changes of the model’s accuracy on other event subsets when it is incrementally trained on event subsets.}
  \label{temporal}
\end{figure*}

\begin{table}[t]
\centering
\renewcommand\arraystretch{1.02}
  \caption{Ablative study of \baby on the SOTA MMD model GAMED across three datasets.}
  \label{ablation}
  \small
  \setlength{\tabcolsep}{5pt}{
  \begin{tabular}{m{1.70cm}m{0.68cm}<{\centering}m{0.68cm}<{\centering}m{0.68cm}<{\centering}m{0.68cm}<{\centering}m{0.68cm}<{\centering}m{0.68cm}<{\centering}}
    \toprule
    \quad \ Methods & Acc. & AUC & F1 & F1$_{\text{real}}$ & F1$_{\text{fake}}$ & Avg.$\downarrow$ \\
    \hline
    \multicolumn{7}{c}{\textbf{Dataset}: \textit{GossipCop} \citep{shu2020fakenewsnet}} \\
    GAMED \citep{shen2025gamed} & 84.28 & 82.37 & 73.84 & 90.78 & 59.61 & \textbf{-1.54} \\
    \rowcolor{lightgrayv} \ + \textbf{\baby} & \textbf{86.21} & \textbf{83.85} & \textbf{76.17} & \textbf{91.73} & \textbf{60.62} & - \\
    \ \ \textit{w/o} DPM & 85.63 & 82.93 & 75.12 & 90.72 & 59.98 & \textbf{-0.84} \\
    \ \ \textit{w/o} Expert $\mathbf{E}_s$ & 85.57 & 82.97 & 75.35 & 91.22 & 59.88 & \textbf{-0.71} \\
    \ \ \textit{w/o} Feature $\mathbf{\hat e}$ & 85.17 & 83.00 & 75.39 & 90.89 & 59.80 & \textbf{-0.86} \\
    
    \hline
    \multicolumn{7}{c}{\textbf{Dataset}: \textit{Weibo} \citep{jin2017multimodal}} \\
    GAMED \citep{shen2025gamed} & 87.29 & 94.46 & 87.24 & 87.48 & 87.98 & \textbf{-2.15} \\
    \rowcolor{lightgrayv} \ + \textbf{\baby} & \textbf{89.93} & \textbf{95.58} & \textbf{89.90} & \textbf{89.42} & \textbf{90.39} & - \\
    \ \ \textit{w/o} DPM & 89.28 & 94.69 & 89.25 & 88.60 & 89.80 & \textbf{-0.72} \\
    \ \ \textit{w/o} Expert $\mathbf{E}_s$ & 89.39 & 94.73 & 89.36 & 88.76 & 89.86 & \textbf{-0.62} \\
    \ \ \textit{w/o} Feature $\mathbf{\hat e}$ & 89.01 & 94.74 & 88.98 & 88.48 & 89.49 & \textbf{-0.91} \\
    
    \hline
    \multicolumn{7}{c}{\textbf{Dataset}: \textit{Twitter} \citep{boididou2018detection}} \\
    GAMED \citep{shen2025gamed} & 61.90 & 64.08 & 60.88 & 59.07 & 65.71 & \textbf{-5.73} \\
    \rowcolor{lightgrayv} \ + \textbf{\baby} & \textbf{67.21} & \textbf{72.18} & \textbf{67.31} & \textbf{62.98} & \textbf{70.63} & - \\
    \ \ \textit{w/o} DPM & 66.16 & 69.63 & 64.92 & 61.23 & 69.61 & \textbf{-1.75} \\
    \ \ \textit{w/o} Expert $\mathbf{E}_s$ & 66.36 & 69.73 & 66.24 & 61.29 & 69.89 & \textbf{-1.36} \\
    \ \ \textit{w/o} Feature $\mathbf{\hat e}$ & 65.53 & 67.31 & 64.94 & 60.50 & 69.38 & \textbf{-2.53} \\
    \bottomrule
  \end{tabular} }
\end{table}

\subsection{Ablation Study}

To investigate the effectiveness of key modules in \baby, we conduct ablation experiments on three continual MMD datasets using the latest GAMED model as the baseline. The ablative results are reported in Table~\ref{ablation}. For clarity, our compared ablation versions are described as follows:
\begin{itemize}
    \item \textbf{w/o Dirichlet Process Mixture (DPM)} removes the Dirichlet process mixture to expand experts dynamically, and fixes the number of data-specific experts at 4;
    \item \textbf{w/o Expert} $\mathbf{E}_s$ removes the data-shared expert optimized using an exponential moving average in the MoE module;
    \item \textbf{w/o Feature} $\mathbf{\hat e}$ eliminates the environment feature $\mathbf{\hat e}$, indirectly removing the environment dynamics module.
\end{itemize}
Generally, the experimental results demonstrate that removing any modules consistently degrades the model’s performance, which demonstrates the effectiveness of the proposed modules in enhancing detection performance. Specifically, the average decline in detection performance for the three ablation versions compared to the full version can be roughly ranked as: Feature $\mathbf{\hat e}$ > DPM > Expert $\mathbf{E}_s$. 
First, the feature $\mathbf{\hat e}$ predicts the future environmental distribution using a continuous-time dynamics system and is supervised by the environmental distribution specified by Gaussian distributions, which are generated for each batched data. Removing this feature deprives it of the ability to predict the environmental distribution of future (test) data, thereby reducing its generalization capability to unseen data. Therefore, this module has the most severe impact on the model’s performance on unseen test sets.
Then, DPM dynamically expands the number of experts during training. When this dynamical process is removed and the number of data-specific experts is fixed at 4, overlapping distributions of training data may cause interference among the parameters of the experts, which exacerbates the model’s catastrophic forgetting problem and negatively impacts its performance.
Finally, the expert $\mathbf{E}_s$ represents knowledge shared across data. When this module is removed, the shared knowledge is dispersed among the data-specific experts, which not only affects training efficiency but also reduces the model’s detection performance.

\begin{figure}[t]
  \centering
  \includegraphics[scale=0.195]{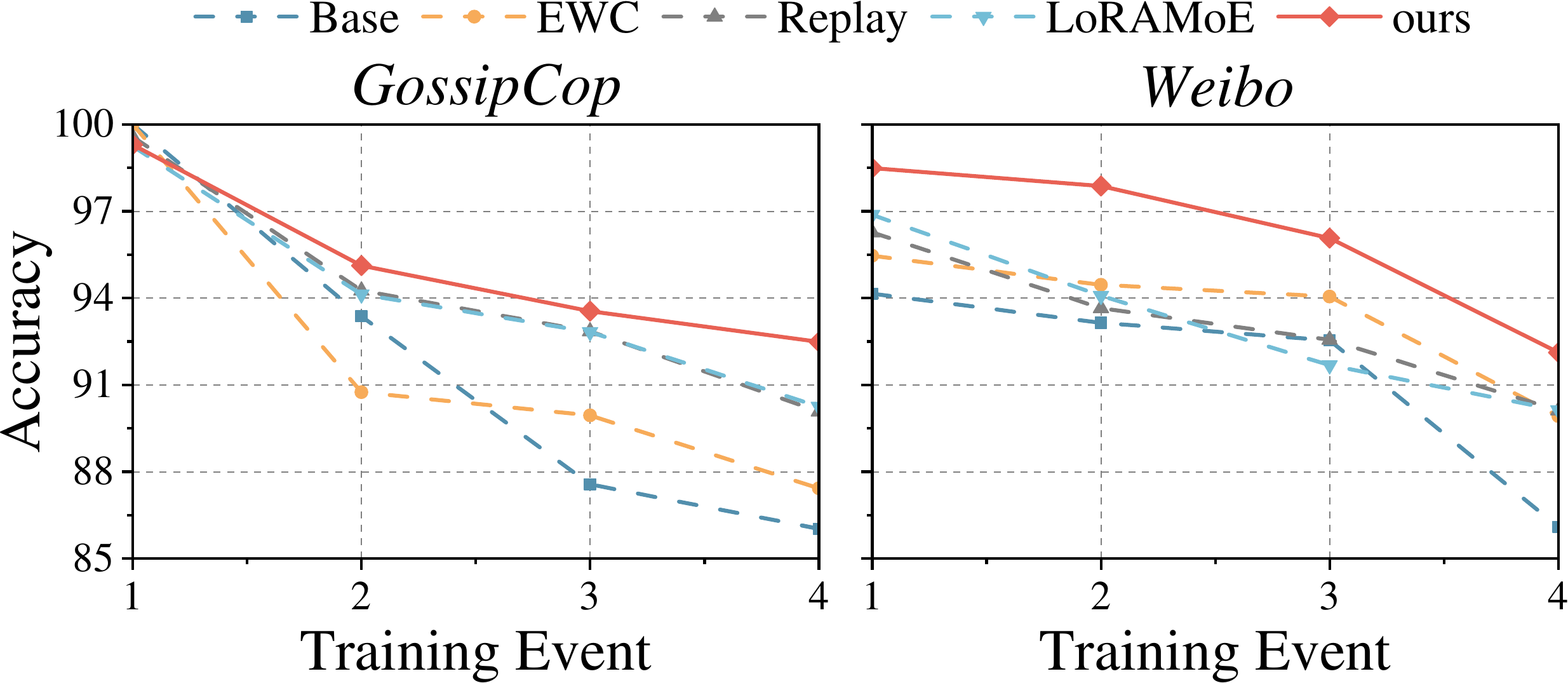}
  \caption{The changes of the model’s accuracy on the first event subsets when it is continually trained on event subsets.}
  \label{forgetting}
\end{figure}

\subsection{Evaluation of Catastrophic Forgetting}

To evaluate whether our method addresses the catastrophic forgetting problem, we present two experimental results, which illustrate the changes of the model’s detection accuracy on other event subsets when it is incrementally trained on event subsets, as shown in Fig.~\ref{temporal}. Additionally, to provide a more concrete visualization of the model’s ability to mitigate catastrophic forgetting, we report the model’s accuracy on the first event subset in Fig.~\ref{forgetting}.

In Fig.~\ref{temporal}, we present the changes in accuracy on other event subsets when the detection model is continually trained using four event subsets. We also compare the performance of our method \baby with prevalent continual learning approaches. Generally, in most scenarios, the model achieves its best performance on a given event immediately after completing training on that event, with performance gradually declining before and after training. However, compared to existing continual learning methods, our method exhibits a lower degree of performance degradation, especially on the first two events.

To more clearly illustrate the decline in the model’s detection performance on the first event during the continual training process, we present Fig.~\ref{forgetting}. It becomes evident that our method consistently outperforms other methods in maintaining performance on the first event during training on subsequent events. For example, on the \textit{GossipCop} dataset, when training reaches the last event, the accuracy of our method on the first event decreases by approximately 6.81. In contrast, the accuracy drops for EWC, Replay, and LoRAMoE are approximately 12.57, 9.49, and 9.00, respectively. This demonstrates that our model is more effective in addressing the catastrophic forgetting problem in the continual MMD task.

\subsection{Sensitivity Analysis}

\begin{figure}[t]
  \centering
  \includegraphics[scale=0.20]{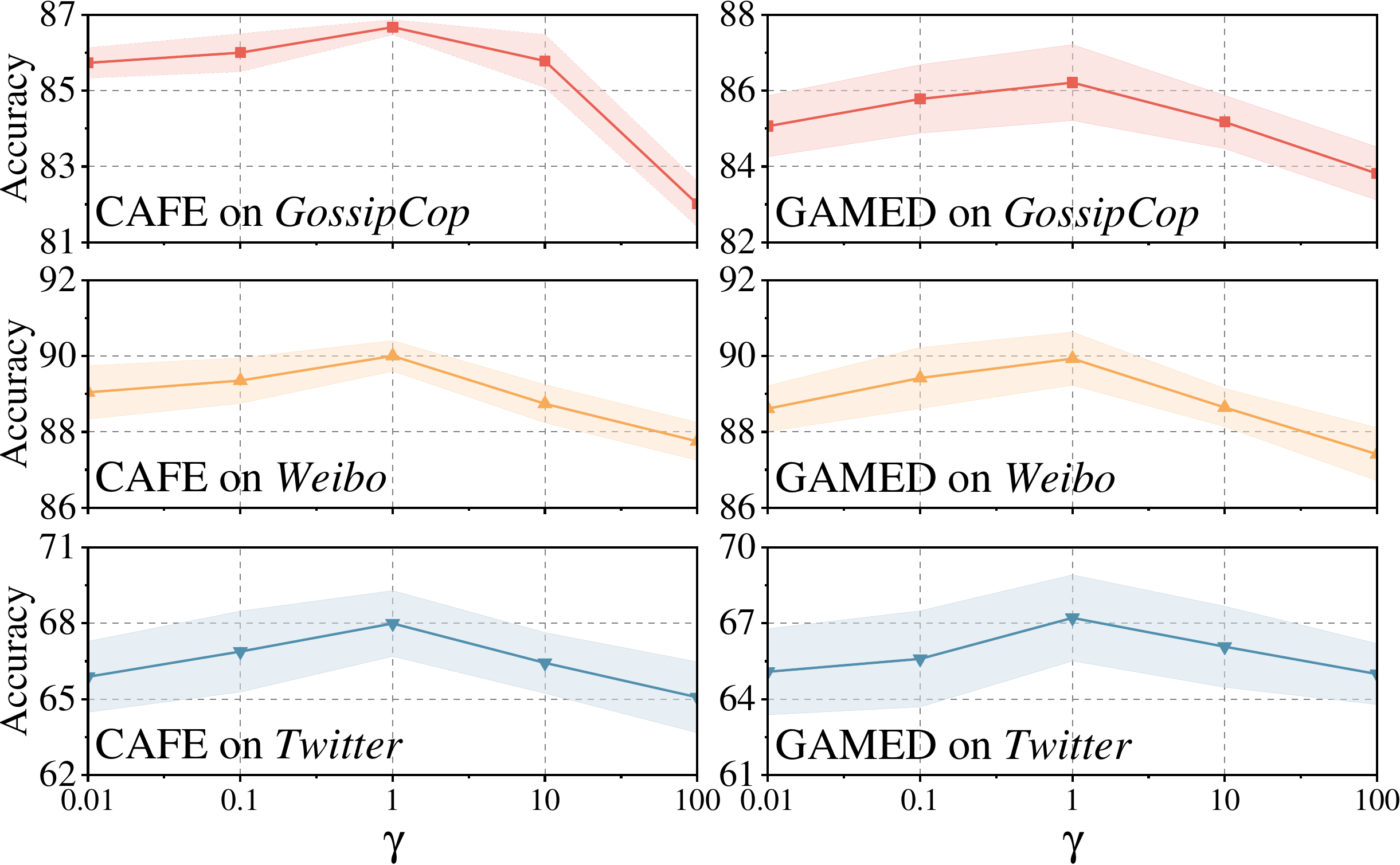}
  \caption{Sensitivity analysis of the parameter $\gamma$.}
  \label{sensitivity}
\end{figure}

To investigate the impact of key parameters in \baby on detection performance, we conduct a sensitivity analysis on a parameter $\gamma$, which is the balancing parameter for the $\mathcal{L}_{DM}$ term in the overall loss function $\mathcal{L}$, across three datasets. The results are shown in Fig.~\ref{sensitivity}. The experimental results demonstrate that existing MMD models, e.g., CAFE and GAMED, achieve optimal accuracy when $\gamma = 1$ on all datasets. Moreover, as $\gamma$ decreases or increases, the model's performance consistently declines. This decline occurs because a lower $\gamma$ value leads to insufficient optimization of the dynamics model, resulting in the sampled feature $\mathbf{\hat e}$ from the predicted distribution of the dynamics model being less discriminative, thereby degrading model performance. Conversely, as $\gamma$ increases, the optimization process overly prioritizes the dynamics model, neglecting the classification model, which also leads to under-optimization of the classification model. These findings provide substantial evidence to support the implementation of \baby.

\begin{figure}[t]
  \centering
  \includegraphics[scale=0.48]{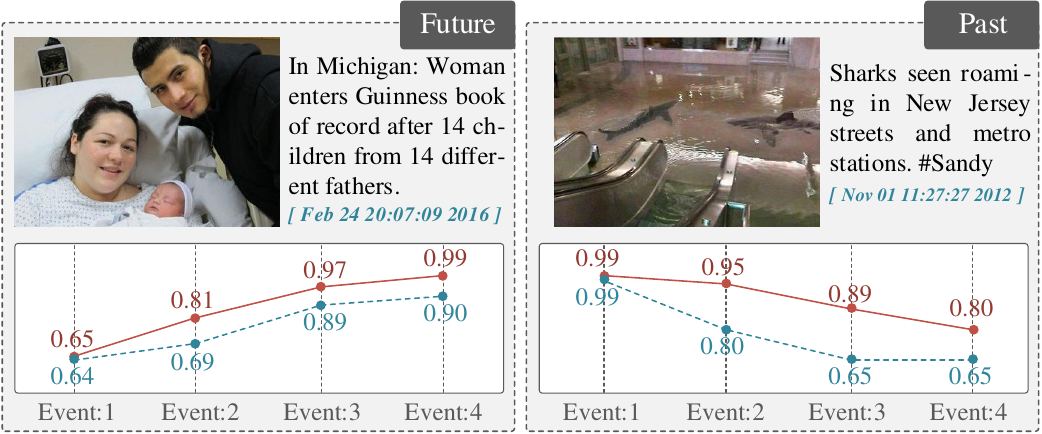}
  \caption{Case study that demonstrates the fake class predicted probabilities during the continual training on 4 events. Red and blue lines are \baby and LoRAMoE, respectively.}
  \label{case}
\end{figure}

\subsection{Case Study}

In Fig.~\ref{case}, we present two representative misinformation examples, one from a future sample in the test set and the other from a past sample in the first training event. Fig.~\ref{case} illustrates the trends in the predicted probability of the fake class for these two examples as the model undergoes continual training. 
Generally, for the future sample, both our method and LoRAMoE exhibit an upward trend, with our method showing a more significant increase. This indicates that our approach can enhance the model's generalization ability on unseen events by predicting future data distributions. For the past sample, both methods show a downward trend due to the issue of knowledge forgetting. However, the decline in our method is slower, demonstrating that our proposed dynamically adapted MoE module is more effective at mitigating catastrophic forgetting.

\section{Conclusion}

In this paper, we aim to learn MMD models in continual data streams. By preliminary observations, we find that current continual MMD models are hindered by two primary challenges: past knowledge forgetting and social environment evolving. To address these issues, we propose a new continual MMD method \baby. Specifically, we design a dynamically adapted MoE model with the Dirichlet process mixture to isolate event-specific knowledge, and learn a dynamics model to predict the environmental distribution of future samples.
Our experimental results can demonstrate that \baby consistently and significantly improves the performance of its baseline MMD models and outperforms prevalent continual learning methods.
In our future work, we will further explore the underlying reasons for changes in the social environment and propose more fine-grained solutions to model these aspects of change.

\section*{Acknowledgement}
We acknowledge support for this project from the National Key R\&D Program of China (No.2021ZD0112501, No.2021ZD0112502), the National Natural Science Foundation of China (No.62276113), and China Postdoctoral Science Foundation (No.2022M721321).

\clearpage
\bibliographystyle{ACM-Reference-Format}
\bibliography{reference}

\appendix

\section{Pseudo Algorithm}

We provide the pseudo algorithm in Alg.~\ref{algorithm} to illustrate the overall training pipeline of the \baby method.

\renewcommand{\algorithmicrequire}{\textbf{Input:}}
\renewcommand{\algorithmicensure}{\textbf{Output:}}
\begin{algorithm}[h]
    \caption{Training pipeline of \baby.}
    \label{algorithm}
    \begin{algorithmic}[1]
    \Require Training continual MMD dataset $\{\mathcal{D}_k\}_{k=1}^K$; training iterations $I$.
    \Ensure An optimal MMD model parameterized by $\boldsymbol{\theta}$; dynamics model parameterized by $\boldsymbol{\phi}$.
    \State Initialize $\boldsymbol{\theta}^t$ and $\boldsymbol{\theta}^v$ with the pre-trained weights;
    \State Initialize data-shared and data-specific experts $\mathbf{E}_s$, $\{\mathbf{E}_m\}_{m=1}^M$;
    \For{$k = 1, 2, \cdots, K$}
        \For{$i = 1, 2, \cdots, I$}
            \State Draw a mini-batch data $\mathcal{B}_i$ from $\mathcal{D}_k$;
            \State Extract unimodal and multimodal features $\mathbf{z}_i^t$, $\mathbf{z}_i^v$, $\mathbf{z}_i$;
            \State Calculate the responsibility $\rho_{i, m}$ with Eq.(4);
            \If{$\arg \min_{m} - \log \rho_{i, m} = M + 1$} 
                \State Add a new data-specific expert $\mathbf{E}_{M+1}$;
            \EndIf
            \State Calculate the variational generation loss $\mathcal{L}_{VG}$;
            \State Obtain the feature $\mathbf{e}_i$ with Eq.(7));
            \State Construct the distribution $\mathbb{P}(\mathbf{\hat e}|y=1)$ with Eq.(10);
            \State Calculate the dynamics model loss $\mathcal{L}_{DM}$ with Eq.(13);
            \State Predict $\mathbb{P}(\mathbf{\hat e}_\tau|y=1)$ with the dynamics model $\boldsymbol{\phi}$;
            \State Sample the environmental feature $\mathbf{\hat e}_i$ from $\mathbb{P}(\mathbf{\hat e}_\tau|y=1)$;
            \State Calculate the veracity prediction loss $\mathcal{L}_{VP}$ with Eq.(2);
            \State Optimize $\boldsymbol{\theta}$ and $\boldsymbol{\phi}$ with $\mathcal{L}$ in Eq.(3).
        \EndFor
    \EndFor
    \end{algorithmic}
\end{algorithm}

\section{Additional Experiments}

In this section, we provide a brief description of the baseline MMD models and continual learning methods in our comparative experiments. Additionally, we conduct a convergence analysis to primarily evaluate whether our dynamics model can correctly converge to learn the dynamic data distribution.

\subsection{Baseline Models}

We select 6 MMD models as baselines to evaluate that our proposed method can enhance their performance in continual MMD scenarios. They are briefly described as follows:
\begin{itemize}
    \item \textbf{Base} \citep{he2016deep,devlin2019bert} directly concatenate semantic features output by text and image encoders, \eg BERT \citep{devlin2019bert} and ResNet \citep{he2016deep}, for veracity classification.
    \item \textbf{SAFE} \citep{zhou2020safe} introduces inter-modal similarity as an additional feature and optimizes modal alignment as a regularization.
    \item \textbf{MCAN} \citep{wu2021multimodal} designs a cross-modal co-attention mechanism to compute the attention features between text and image modalities, and frequency images.
    \item \textbf{CAFE} \citep{chen2022cross} employs a variational distribution to measure modality inconsistency and uses this inconsistency as the weight for unimodal and multimodal feature fusion.
    \item \textbf{BMR} \citep{ying2023bootstrapping} designs an improved mixture-of-experts model to capture multimodal multi-view article features.
    \item \textbf{GAMED} \citep{shen2025gamed} incorporates the AdaIN technique \citep{huang2017arbitrary} into the BMR model to perform progressive feature refinement and dynamic feature weighting.
\end{itemize}
Additionally, we compare three prevalent continual learning methods as follows:
\begin{itemize}
    \item \textbf{EWC} \citep{kirkpatrick2016overcoming,han2021continual} quantifies the importance of different model parameters and prioritizes to protect ones that significantly impact the performance of previous tasks, thereby preserving critical knowledge from old tasks. \citet{han2021continual} directly uses EWC to alleviate the forgetting in text-only misinformation detection.
    \item \textbf{Replay} \citep{robins1995catastrophic,lee2021dynamically} maintains memory of old tasks by storing a portion of previous data, which are then mixed with new task training to preserve knowledge of prior tasks. According previous works \citep{lopez2017gradient} and our empirical results, we fix the rate of replay data to 25\%.
    \item \textbf{LoRAMoE} \citep{dou2024loramoe} trains separate low-rank experts for different tasks, thereby ensuring that the learning of one task does not interfere with others.
\end{itemize}

\subsection{Computation Budgets}

Since \baby introduces an additional MoE system and incorporates an extra training process for the dynamics model, it is essential to investigate the time consumption of \baby. Accordingly, we compare the training time of existing continual learning methods with that of \baby. Table~\ref{time} shows the relative time consumption of each method compared to the base model, which does not employ any continual learning strategies. Specifically, from Table~\ref{time}, we observe that existing continual learning methods such as EWC, Replay, and LoRAMoE all exhibit higher time consumption than the base model, with the highest increase reaching 1.13 times that of the base model. Among these methods, the replay method incurs the most significant time overhead due to the need to store historical data and retrain on it, which directly increases the training time. In contrast, our model \baby demonstrates lower time consumption than the base model on the \textit{Weibo} and \textit{Twitter} datasets. This is because our training process employs an early stopping strategy to control the number of training iterations. The reduced training time further demonstrates that our method converges faster than the base model while achieving superior detection performance.

\begin{table}[t]
\centering
\renewcommand\arraystretch{1.0}
  \caption{Time complex analysis of various continual methods.}
  \label{time}
  \small
  \setlength{\tabcolsep}{5pt}{
  \begin{tabular}{m{1.05cm}<{\centering}m{0.90cm}<{\centering}m{0.90cm}<{\centering}m{1.0cm}<{\centering}m{1.10cm}<{\centering}m{1.10cm}<{\centering}}
    \toprule
    Dataset & Base & EWC & Replay & LoRAMoE & \baby \\
    \hline
    \textit{GossipCop} & 1$\times$ & 1.09$\times$ & 1.07$\times$ & 1.03$\times$ & 1.08$\times$ \\
    \textit{Weibo} & 1$\times$ & 1.06$\times$ & 1.13$\times$ & 1.06$\times$ & 0.97$\times$ \\
    \textit{Twitter} & 1$\times$ & 1.11$\times$ & 1.09$\times$ & 1.02$\times$ & 0.96$\times$ \\
    \bottomrule
  \end{tabular} }
\end{table}


\end{document}